\documentclass[preprint,12pt]{elsarticle}

\usepackage{amssymb}
\usepackage{amsmath}

\usepackage{xcolor}

\usepackage{multirow}
\usepackage{booktabs}

\journal{Neurocomputing}

\begin{document}

\begin{frontmatter}
\makeatletter
\def\ps@pprintTitle{%
 \let\@oddhead\@empty
 \let\@evenhead\@empty
\def\@oddfoot{\hfill\footnotesize \textit{This manuscript has been accepted for publication in Neurocomputing (Elsevier).} \hfill}
 \let\@evenfoot\@oddfoot}
\makeatother




\title{What Are We Really Measuring? Rethinking Dataset Bias in Web-Scale Natural Image Collections via Unsupervised Semantic Clustering}

\author[first]{Amir Hossein Saleknia\corref{cor1}}
\cortext[cor1]{Corresponding author}

\affiliation[first]{organization={School of Electrical Engineering, Iran University of Science and Technology (IUST)},
            city={Tehran},
            country={Iran}}
            
\ead{a\_saleknia@alumni.iust.ac.ir}

\author[second]{Mohammad Sabokrou}

\affiliation[second]{
            organization={Okinawa Institute of Science and Technology (OIST)},
            city={Okinawa},
            country={Japan}}

\ead{mohammad.sabokrou@oist.jp}

\begin{abstract}

In computer vision, a prevailing method for quantifying dataset bias is to train a model to distinguish between datasets. High classification accuracy is then interpreted as evidence of meaningful semantic differences. This approach assumes that standard image augmentations successfully suppress low-level, non-semantic cues, and that any remaining performance must therefore reflect true semantic divergence. We demonstrate that this fundamental assumption is flawed within the domain of large-scale natural image collections. High classification accuracy is often driven by resolution-based artifacts, which are structural fingerprints arising from native image resolution distributions and interpolation effects during resizing. These artifacts form robust, dataset-specific signatures that persist despite conventional image corruptions. Through controlled experiments, we show that models achieve strong dataset classification even on non-semantic, procedurally generated images, proving their reliance on superficial cues. To address this issue, we revisit this decades-old idea of dataset separability, but not with supervised classification. Instead, we introduce an unsupervised approach that measures true semantic separability. Our framework directly assesses semantic similarity by clustering semantically-rich features from foundational vision models, deliberately bypassing supervised classification on dataset labels. When applied to major web-scale datasets, the primary focus of this work, the high separability reported by supervised methods largely vanishes, with clustering accuracy dropping to near-chance levels. This reveals that conventional classification-based evaluation systematically overstates semantic bias by an overwhelming margin. Our findings offer a modern re-evaluation of how dataset separability is interpreted, suggesting that unsupervised clustering provides a more reliable measure of semantic bias in large-scale natural image collections than supervised classification.

\end{abstract}



\begin{keyword}
Dataset Bias \sep Resolution Artifacts \sep Semantic Clustering \sep Bias Measurement \sep Representation Learning \sep Foundation Models
\end{keyword}

\end{frontmatter}



\section{Introduction}
\label{sec1}

This work focuses on dataset bias measurement within web-scale natural image collections. The performance of modern computer vision models is inextricably linked to the data on which they are trained. Consequently, understanding and measuring the biases inherent in these datasets is a problem of fundamental importance. A landmark study by ~\citet{torralba2011unbiased} crystallized this issue by posing a foundational question: do our datasets represent the visual world, or have they become mere instruments for gaming benchmarks? They hypothesized that if datasets were truly unbiased samples from the visual world, it should be difficult to distinguish between them. In this view, an ideal dataset would be indistinguishable from another if both were unbiased samples of the same underlying distribution. To test this assumption, \citet{torralba2011unbiased} introduced a diagnostic task they called \emph{``Name That Dataset.''} They trained a Support Vector Machine (SVM) classifier on low-level image descriptors extracted from 12 widely used vision datasets. Surprisingly, the classifier achieved significantly above‐chance accuracy, which they interpreted as strong evidence of dataset bias driven by dataset‐specific, learnable signatures. They attributed this bias to divergent dataset design choices, including differences in scene content, photographic style, and object framing. Based on these insights, they proposed guidelines for constructing future large-scale datasets with reduced inherent bias.

\begin{figure*}[h]
\centering
\includegraphics[width=1.0\textwidth]{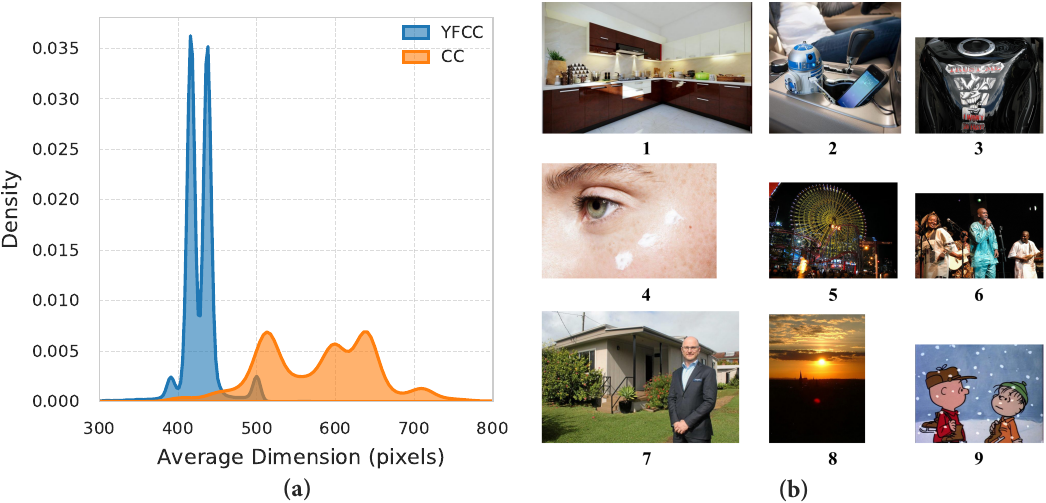} 
\caption{The \emph{"Name That Dataset"} game \citep{torralba2011unbiased} aware of resolution distributions. These images are sampled from the YFCC and CC datasets. We maintain their relative sizes. Additionally, we provide a plot showing the resolution distribution of the training samples from each dataset. \emph{Can you guess which dataset each image is from?} Once resolution effects are considered, dataset separability becomes significantly more apparent. \small{(Answer: YFCC: 3, 5, 6, 8, 9; CC: 1, 2, 4, 7)}}
\label{figure-1}
\end{figure*}

Recently, \citet{liu2024decade} revisited the \emph{``Name That Dataset''} experiment using modern large-scale datasets and powerful deep learning models. They constructed a three-way classification task (YCD) comprising three major web-scale image datasets: YFCC \citep{yfcc100m}, CC \citep{CC}, and DataComp \citep{DataComp}. Their results showed that their baseline model could accurately identify the source dataset of an image, achieving 84.7\% accuracy on this task. While these findings were presented as surprising, we identify a critical oversight: their narrow focus on classification performance obscures the nature of the underlying signals. This limitation becomes apparent when contrasting their separate findings: their user study reported that 20 expert humans achieved a mean accuracy of only 45.4\% on the YCD task, while their experiments showed that a minimal ConvNeXt  \citep{ConvNeXt} variant with just 7,000 parameters could achieve 72.4\% accuracy. This paradox, where a simple model surpasses humans on a task that ostensibly requires semantic understanding, motivated us to investigate what the model was actually learning. Their assumption that models surviving standard corruptions such as color jittering and Gaussian noise must be using high-level semantics is only valid if such corruptions disrupt all low-level artifacts. However, they concentrated on easily disruptable cues like color statistics, overlooking more persistent structural artifacts. Our investigation began with a simple observation of the raw images: a noticeable variation in native resolutions across datasets. We systematically measured this, defining resolution proxy as the average of an image's height and width. Visualization via Kernel Density Estimation (KDE) revealed clear, distinct resolution distributions for different datasets, as shown in Figure~\ref{figure-1} (a). This finding suggests that even when images are uniformly resized for training, the resizing process can introduce dataset-specific interpolation artifacts: subtle yet consistent cues that a model may learn to exploit. As a result, high classification accuracy may reflect reliance on these robust but superficial resolution-based signals rather than true semantic understanding, revealing a fundamental flaw in the conventional interpretation of dataset separability.

Building upon this insight, we revisited the \emph{``Name That Dataset''} game, this time with an explicit awareness of resolution distributions and a commitment to preserving images' relative sizes. As Figure~\ref{figure-1} (b) illustrates, once these resolution patterns are made visible, distinguishing between datasets becomes almost trivial. What was previously treated as a challenging classification problem now reduces to a relatively simple visual inspection task, where even a brief glance can reveal the dataset source. This highlights how low-level, often overlooked features like native image resolution can serve as strong, dataset-specific fingerprints.

Through controlled experiments, we show that deep models exploit resolution artifacts, even when all images are uniformly resized. These low-level cues alone can lead to strong dataset classification performance. This presents a critical risk: high accuracy may be misleading if models rely on superficial artifacts rather than meaningful semantic differences. Throughout this paper, we use the term ``semantic bias'' to refer specifically to the degree to which images from different web-scale natural image collections can be distinguished based solely on their high-level visual content (objects, scenes, actions, compositional themes), independent of low-level acquisition and processing artifacts such as resolution, JPEG compression, or color statistics.

We revisit the core idea proposed by \citet{torralba2011unbiased}, that dataset separability can be interpreted as a measure of dataset bias, but introduce a new style of measurement designed to avoid accounting for trivial signals. In doing so, we demonstrate that dataset bias in large web-scale natural images is far less pronounced than suggested by recent work such as \citet{liu2024decade}.

Specifically, to address the limitations of supervised classification for semantic bias assessment, we propose an unsupervised, artifact-agnostic approach. Our method leverages semantically rich embeddings from DINOv2 \citep{DinoV2}, clustering images based on content rather than superficial signals. This reveals the intrinsic semantic structure of datasets while minimizing sensitivity to low-level artifacts. Our work makes three key contributions in the context of web-scale natural image datasets: (1) We demonstrate that these datasets exhibit distinct resolution profiles. Even after resizing to a common dimension, the interpolation artifacts derived from these profiles provide a persistent, non-semantic signal that can be exploited for dataset classification; (2) we provide comprehensive evidence that these artifacts alone can drive high dataset classification performance for such datasets, potentially misleading bias assessments; and (3) we introduce an unsupervised evaluation framework that directly measures semantic bias in large-scale natural image collections, offering a re-evaluation of how dataset separability can be interpreted within this domain.

The remainder of this paper is organized as follows. Section~\ref{sec2} reviews related work on dataset bias and its measurement. In Section~\ref{sec3}, we formalize the problem and introduce our proposed unsupervised framework for semantic bias assessment. Section~\ref{sec4} presents our core experimental findings, demonstrating that resolution artifacts alone can drive high performance in supervised dataset classification. Section~\ref{sec5} then evaluates our unsupervised method, showing it reveals significantly lower semantic bias and is robust to scaling and architectural changes. Section~\ref{sec6} provides a qualitative and quantitative characterization of the residual semantic bias that persists once low-level artifacts are controlled for. Finally, Section~\ref{sec7} discusses the broader implications of our findings, revisits prior interpretations in light of our evidence, and outlines limitations and future directions before concluding in Section~\ref{sec8}.

\section{Related Works}
\label{sec2}
Deep Neural Networks (DNNs) have achieved remarkable success across numerous tasks, but their progress is hindered by dataset bias, which negatively impacts model generalization and fairness. Early work by \citet{torralba2011unbiased} showed that simple classifiers could distinguish between datasets using image descriptors such as Histogram of Oriented Gradients (HOG) \citep{HOG} and GIST \citep{gist}, revealing hidden dataset-specific signals. Over the years, building upon the diagnostic framework introduced by them, researchers have leveraged dataset classification as a straightforward tool to evaluate and understand dataset bias across a variety of tasks. For example, \citet{Dataset_bias_emotions} applied three tests, including dataset classification, to identify, illustrate, and measure dataset bias in emotion recognition datasets. Similarly, \citet{Deflating_Dataset_Bias_Using_Synthetic_Data_Augmentation} identified and quantified dataset bias in vision datasets for autonomous vehicles using confusion matrices of models trained for dataset classification, and addressed the issue through synthetic data augmentation. \citet{Dataset_bias_plate_recognition} employed dataset classification to measure bias across License Plate Recognition (LPR) datasets. To investigate the degree and cause of dataset bias in the Visual Question Answering (VQA) task, \citet{Dataset_bias_VQA} performed dataset classification on joint features derived from Image–Question–Answer (IQA) triplets.

Beyond its initial application in natural images, dataset classification has become a widely adopted tool for measuring dataset bias across diverse domains. In medical imaging, \citet{Medical} demonstrated that brain MRI scans’ dataset membership could be identified with 71.5\% accuracy across multi-site neuroimaging datasets, revealing structural biases. They further showed that harmonization across imaging sites reduces this bias while preserving biological variability. Similarly, \citet{Dataset_bias_medical} employed dataset classification to evaluate dataset bias across 15 large-scale public datasets of T1-weighted brain MRI scans. The approach has also been extended beyond computer vision; for example, in Natural Language Processing (NLP), dataset classification has been employed to analyze dataset bias \citep{NLP_2025}.

Despite persistent efforts, dataset bias remains a significant challenge across the machine learning pipeline. Early work by \citet{UndoingDatasetBias2013} showed that explicitly modeling and removing dataset-specific biases can improve cross-dataset generalization. The rise of deep learning further revealed that even powerful representations (DeCAF \citep{DeCAF2013}) can capture and amplify dataset artifacts \citep{DeCAF2015}, ultimately hindering generalization even as overall recognition performance improves. This problem has spurred a wide array of mitigation strategies. Tools like REVISE \citep{revise2020} facilitate the automatic detection of various biases, enabling systematic analysis. Several methods operate directly on the data distribution. For example, REPAIR \citep{cross_dataset_gen_0} formulates bias reduction as a minimax optimization problem, adversarially learning instance-specific resampling weights. Similarly, AFLite \citep{aflite2020} adversarially filters a dataset by iteratively pruning the examples most easily predicted by a probe model, thereby systematically removing instances solvable via spurious correlations. Another category of approaches intervenes in the learning process itself. For instance \citet{balanced_not_enough2019} use adversarial training to strip protected attributes such as gender and race from a model's intermediate representations. There are also gradient-based methods like PGD \citep{PGD2023} which automatically identifies \textit{bias-conflicting samples} \citep{LearningFromFailure2020} via their per-sample gradient norms. PGD's core contribution is to debias a model by oversampling these high-norm samples, achieving robust performance without explicit bias supervision.

However, the diagnostic utility of all these efforts, whether for identifying, measuring, or mitigating bias, rests on a critical and largely unexamined assumption: that high accuracy in supervised dataset classification reliably reflects meaningful semantic bias. Recent evidence from \citet{clip_traces} undermines this assumption, showing that visual encoders automatically encode metadata related to image processing (e.g., JPEG compression) and acquisition (e.g., camera model) alongside semantic information. Consequently, the metric itself is corrupted by superficial artifacts, potentially rendering subsequent bias assessments futile. This risk is evident in works such as \citep{Understanding_Bias}, which explicitly interprets high dataset classification accuracy as direct evidence of semantic bias and seeks to characterize its supposed forms. Our work challenges this premise by demonstrating that the measured separability often stems not from semantic divergence, but from exploitable low-level artifacts, necessitating a fundamental rethinking of how dataset bias is assessed.

\section{Proposed Method: Unsupervised Semantic Bias Assessment}
\label{sec3}

The so-called \emph{``Name That Dataset''} task appears deceptively straightforward. Given samples \(x\) drawn from \(K\) datasets, indexed by \(D \in \{1, \dots, K\}\), a standard approach is to train a \(K\)-way classifier to predict the dataset label \(D\):

\begin{equation}
g: X \;\longrightarrow\; \{1,\dots,K\}
\end{equation}

and evaluate its performance via classification accuracy:

\begin{equation}
\alpha \;=\; \mathbb{P}\bigl(g(X) = D\bigr).
\end{equation}

If \(\alpha\) significantly exceeds random chance \((1/K)\), one typically concludes that the datasets are distinguishable, or in other words, that they exhibit \emph{``bias''} relative to one another.


Formally, we model each data point \(X\) as a pair \((S, N)\), where \(S\) is the semantic signal and \(N\) the nuisance component. A Bayes-optimal classifier would use the full joint distribution:

\begin{equation}
p_i(s,n) = \mathbb{P}(S = s,\, N = n \mid D = i)
\end{equation}

and predict using:

\begin{equation}
g^*(s,n) = \arg\max_{i} \; p_i(s,n).
\end{equation}

However, our objective is not to model superficial artifacts, but rather to assess differences that arise purely from semantic content. Ideally, this would involve classification based on the marginal distribution over \(S\):

\begin{equation}
\begin{aligned}
g^*_S(s)
&= \arg\max_{i} \, \mathbb{P}(D = i \mid S = s) \\
&= \arg\max_{i} \int p_i(s, n) \, \mathrm{d}n \\
&= \arg\max_{i} \, p_i^S(s)
\end{aligned}
\end{equation}

where:

\begin{equation}
p_i^S(s) = \mathbb{P}(S = s \mid D = i).
\end{equation}

In other words, meaningful semantic bias corresponds to how well one can distinguish datasets using \(S\) alone, ignoring all information from the nuisance variables \(N\).

Since $N$ is not directly observable, we cannot implement $g^*_{S}$ in closed form. To address this, we propose an alternative method for assessing semantic bias that avoids training a classifier directly on dataset labels. Instead, we leverage semantically rich, general-purpose visual representations extracted from a foundational self-supervised model such as DINOv2. 

Recall from our formal definition in Section~3 that semantic bias refers to the degree to which images from different datasets can be distinguished solely based on their high-level visual content. This definition naturally leads to a practical question: what feature space should we use to measure such semantic distinctions? Two key insights motivate our choice of DINOv2.

First, consider what dataset bias fundamentally represents: it is ultimately about measuring the similarities and differences between images across different collections. If two datasets are semantically similar, their images should cluster together; if they are semantically distinct, they should form separate clusters. Therefore, a feature space that organizes images by semantic similarity is intrinsically suitable for bias assessment.

DINOv2 provides exactly such a space. The model is fundamentally trained to be invariant to the very resolution artifacts we identified as the primary drivers of supervised dataset classification. It learns by enforcing that different visual transformations of the same image must produce identical feature representations. This means that whether the model sees a zoomed-in paw, a wide shot of the whole dog, a low-resolution thumbnail, or a color-adjusted version, the output vector must remain the same. Critically, these transformations include random resized crops at varying scales and aspect ratios, which explicitly forces the model to strip away resolution-based fingerprints and interpolation artifacts, extracting only the invariant semantic content that persists across these transformations. Therefore, this loss function makes DINOv2 features substantially less sensitive to low-level artifacts. This aligns with recent empirical evidence suggesting that self-supervised models can be less sensitive to low-level metadata traces compared to contrastive vision-language or supervised baselines \citep{clip_traces}.

Second, DINOv2 is trained on LVD-142M \citep{DinoV2}, a diverse and carefully curated dataset of 142 million images drawn from the same domain as our target datasets: web-scale natural images. LVD-142M is specifically designed to cover a wide range of visual concepts, objects, scenes, and compositional themes representative of web-scale imagery. Consequently, the feature space encodes the similarities and differences present in this distribution. Since our target datasets (YFCC, CC, DataComp, WIT, LAION) are also web-scale natural image collections, we expect DINOv2 to capture the meaningful semantic variations that exist among them, including differences in object categories, scene types, and compositional styles commonly found in web-scale images. However, we acknowledge that semantic differences falling outside the distribution of LVD-142M, such as highly specialized visual concepts or fine-grained distinctions not well-represented in the training data, may remain partially or fully invisible to our analysis. This domain alignment makes DINOv2 a well-suited proxy for measuring semantic relationships within web-scale natural image collections, while being potentially limited for out-of-distribution concepts.

Together, these training objectives push the model to discard superficial pixel variations and discover what truly defines visual concepts versus accidental details like background or lighting. Regularization techniques prevent the model from collapsing all images into the same output, ensuring it maintains a rich and structured feature space where semantically similar content clusters together based on high-level concepts. The result is a learned visual feature space where distance strongly correlates with semantic meaning: images of dogs tend to cluster closer to each other than to cats, and objects with similar functions often group together. The feature space exhibits a hierarchical structure, with different dog breeds forming a cluster that sits nearer to wolves than to birds, while vehicles like cars and trucks share a distinct neighborhood. Importantly, this organization extends beyond object-centric concepts to scene-level semantics: images of beaches, forests, and cityscapes also form coherent clusters, demonstrating that DINOv2 captures both object and scene semantics simultaneously. This multi-level semantic organization will be empirically examined in Section~\ref{sec6}, where we analyze the residual clustering patterns across datasets to reveal how these different semantic notions manifest in practice.

\begin{figure*}[t]
\centering
\includegraphics[width=0.8\textwidth]{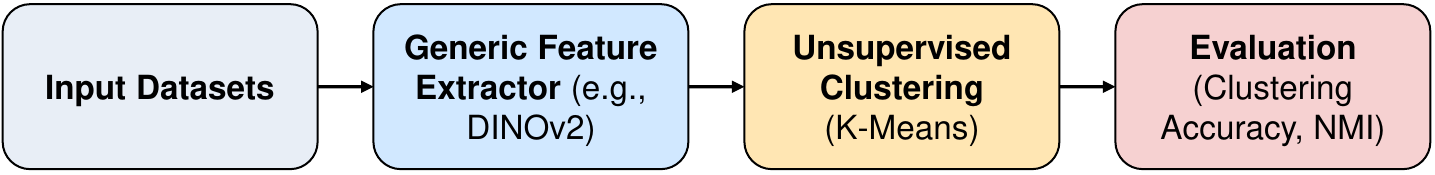} 
\caption{The overview of our proposed unsupervised semantic bias assessment pipeline. Images from multiple datasets are embedded via a generic pretrained feature extractor (e.g., DINOv2), clustered using K-Means, and evaluated using clustering accuracy and normalized mutual information (NMI). Dimensionality reduction via UMAP is applied as an auxiliary step for computational tractability before clustering. See~\ref{app:robustness} for comprehensive robustness analysis.}
\label{figure-2}
\end{figure*}

We emphasize that these embeddings, while powerful, remain learned representations shaped by the model's training data and objectives; they serve as a practical approximation of \(g^*_S\) rather than an objective ground truth for semantics. An overview of our proposed pipeline is illustrated in Figure~\ref{figure-2}.

Let \(\{x_i, D_i\}_{i=1}^n\) be our dataset, where each image \(x_i\) has a corresponding dataset label \(D_i \in \{1, \dots, K\}\). We first extract DINOv2 embeddings \(\Phi_{\mathrm{DINO}}(x_i)\), which are high-dimensional feature vectors. To mitigate the curse of dimensionality and make the subsequent clustering more tractable, we apply the UMAP \citep{UMAP} dimensionality reduction method to project these features into a lower-dimensional space \(\mathbb{R}^d\). Crucially, we choose UMAP over linear alternatives like PCA because of its demonstrated ability to better preserve the global semantic structure and manifold topology of high-dimensional data \citep{UMAP}, which is essential for maintaining meaningful relationships between images in our reduced feature space.

\begin{equation}
\begin{aligned}
\Phi_i &= \Phi_{\mathrm{DINO}}(x_i) \in \mathbb{R}^D \\
z_i &= \mathrm{UMAP}(\Phi_i) \in \mathbb{R}^{d}
\end{aligned}
\end{equation}

Next, we perform \(K\)-means clustering on the reduced features \(\{z_i\}\), yielding centroids \(\{\mu_k\}_{k=1}^K\) and cluster assignments:

\begin{equation}
\hat{c}_i = \arg\min_{k \in \{1,\dots,K\}} \|z_i - \mu_k\|^2.
\end{equation}

We treat the ground truth dataset labels \(D_i\) as a reference clustering \(c_i = D_i\). To quantify the semantic bias, we measure the similarity between the predicted clustering \(\hat{C} = (\hat{c}_i)\) and this ground truth \(C = (c_i)\) using two complementary metrics.

First, we compute the \emph{clustering accuracy}, defined as the maximum accuracy achieved by optimally matching cluster labels to dataset labels:

\begin{equation}
\text{Acc} = \max_{\pi} \frac{1}{N} \sum_{i=1}^{N} \mathbf{1}(c_i = \pi(\hat{c}_i))
\end{equation}

where $\pi$ is a mapping that assigns each cluster to a unique dataset label. This optimal matching is computed using the \emph{Hungarian algorithm}~\citep{Hungarian}, which finds the alignment that maximizes the number of agreements between the cluster assignments and the true labels.

Second, we report the \emph{Normalized Mutual Information} (NMI), which measures the mutual dependence between the two clusterings, normalized by their entropies:
\begin{equation}
\text{NMI}(C, \hat{C}) = \frac{2 I(C, \hat{C})}{H(C) + H(\hat{C})}
\end{equation}
where \(I(C, \hat{C})\) is the mutual information and \(H(\cdot)\) is the entropy. We report NMI values on a percentage scale from 0 to 100. This is calculated by multiplying the standard NMI value (which ranges from 0 to 1) by 100. A value of 0 indicates independent clusterings, while a value of 100 indicates perfect correlation. Unlike clustering accuracy, NMI is invariant to permutations of the cluster labels, providing a complementary measure of clustering quality. Low values for both metrics indicate that the datasets are not easily separable based on semantic content alone.

In the following sections, we validate this approach through extensive experiments. First, we demonstrate that resolution artifacts alone can drive high dataset classification performance (Section~\ref{sec4}). Then, we show that our unsupervised method reveals significantly lower semantic bias than supervised approaches suggest (Section~\ref{sec5}).

\section{Experiments: Resolution Artifacts in Dataset Classification}
\label{sec4}

This section investigates dataset classification in two phases. First, we confirm that deep networks achieve high accuracy in dataset classification, replicating prior work, and notably find that models from diverse architectural paradigms yield highly similar confusion matrices, which suggests convergence on a common, latent signal. Second, through controlled experiments, we demonstrate this signal stems not from semantics, but from low-level resolution cues. This systematic deconstruction challenges the interpretation of classification accuracy as reliable evidence of semantic bias.

\subsection{Datasets}

We probe dataset biases through an N-way classification task over five large-scale image sources. 
Based on YFCC, CC, DataComp, WIT~\citep{WIT}, and LAION~\citep{Laion}, we construct two image sets. 
The YCD image set combines YFCC, CC, and DataComp and serves as the core configuration for most of our experiments. The YCDLW image set extends YCD by including WIT and LAION, resulting in a five-dataset configuration. From each dataset, we randomly select 10{,}000 images for training and 15{,}000 images for testing.

\subsection{Implementation Details}

In all experiments, we employ a consistent training setup. To accelerate the training process, our models are initialized with ImageNet pre-trained weights and trained for 30 epochs using Stochastic Gradient Descent (SGD) with a batch size of 64, momentum of 0.9, and weight decay of 1e-4. We utilize a polynomial learning rate schedule (polyLR) with an initial learning rate of 0.01 and a power of 0.9. Input images are resized to 224×224 pixels and normalized using ImageNet statistics. Our data augmentation pipeline includes random horizontal flipping, random cropping, color jittering, random grayscale conversion, and Gaussian blur. All implementations are in PyTorch and experiments are executed on an NVIDIA A100 GPU with 80GB memory.

\subsection{Dataset Classification: What’s Happening Behind the Scenes}

We investigate the latent cues used for dataset classification by examining models from fundamentally different architectural families. Specifically, we train and analyze three modern networks: ConvNeXt V2-Tiny \citep{ConvNeXt} as a representative of modern convolutional architectures, MViTv2-Tiny \citep{MViTv2} for its multiscale vision transformer approach, and EfficientVit-B3 \citep{EfficientVit} as a hybrid model. These architectures differ substantially in both structure and inductive biases, allowing us to evaluate whether their behaviors converge on the same latent signals within the data or focus on different clues based on their specifications.

\begin{figure*}[h]
\centering
\includegraphics[width=1.0\textwidth]{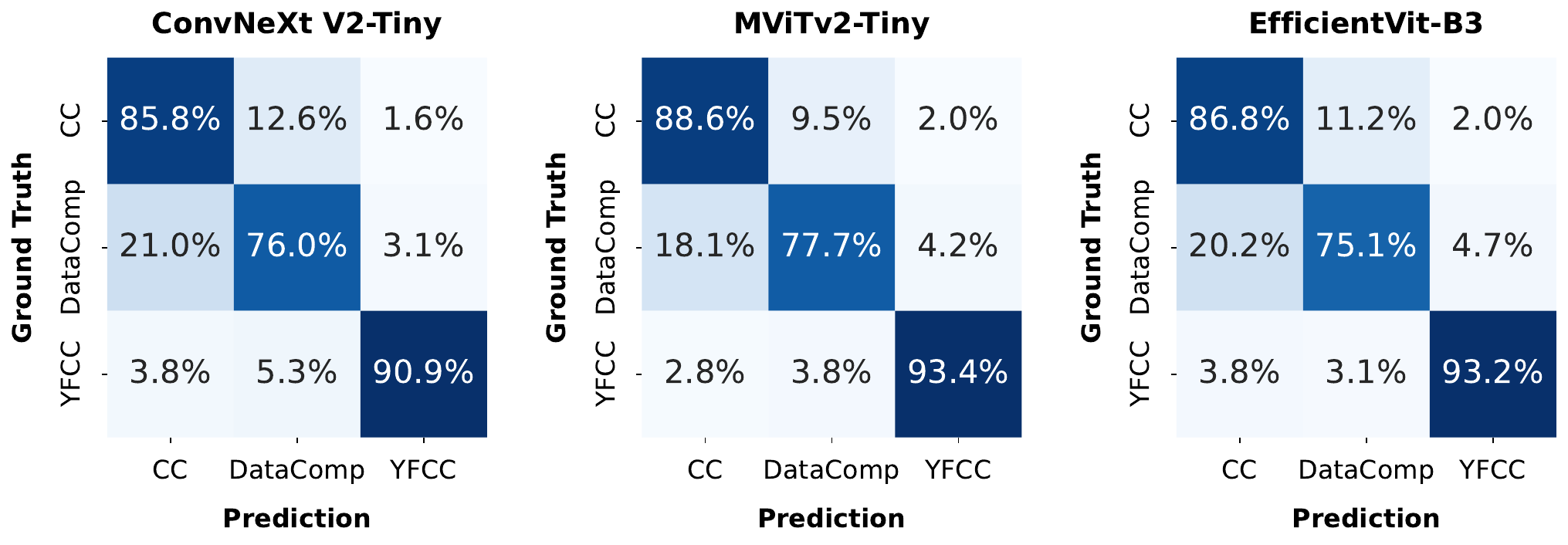} 
\caption{Normalized confusion matrices (per-class percentage) for three models evaluated on YCD. Each matrix shows the performance of: (left) ConvNeXt V2-Tiny, (center) MViTv2-Tiny, and (right) EfficientVit-B3.}
\label{figure-3}
\end{figure*}

\begin{figure*}[h]
\centering
\includegraphics[width=0.9\textwidth]{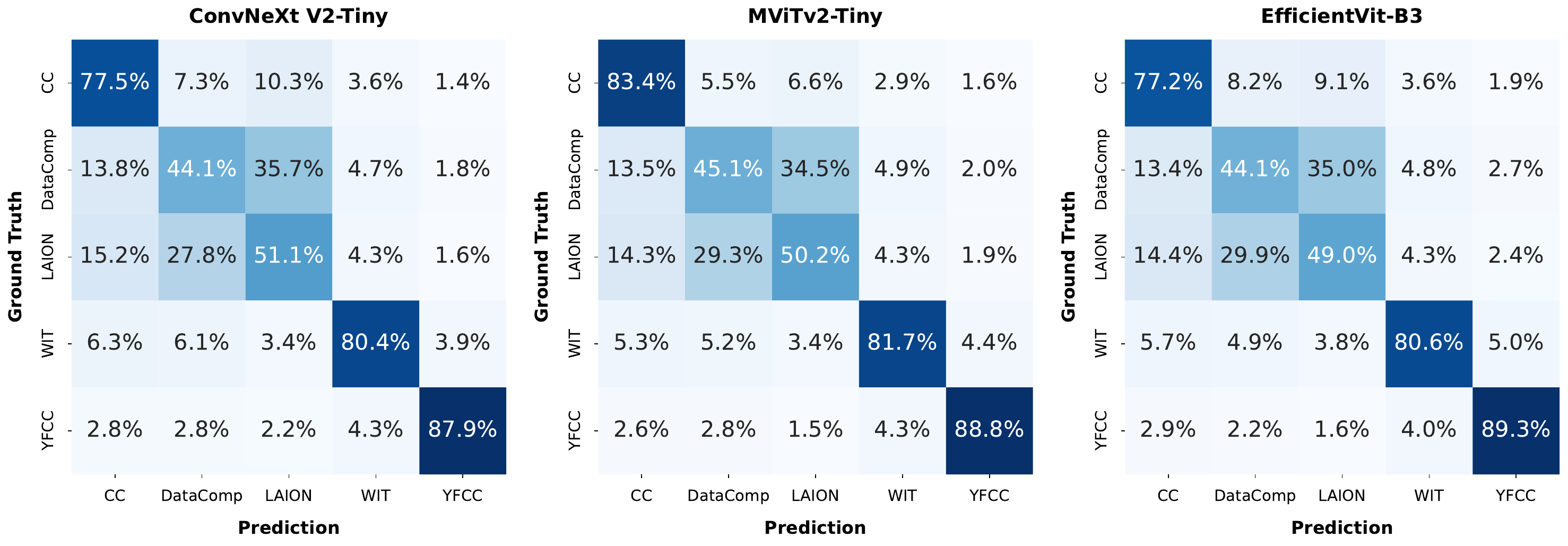} 
\caption{Normalized confusion matrices (per-class percentage) for three models evaluated on YCDLW. Each matrix shows the performance of: (left) ConvNeXt V2-Tiny, (center) MViTv2-Tiny, and (right) EfficientVit-B3.}
\label{figure-4}
\end{figure*}

The classification results for the YCD and YCDLW image sets are presented in Table~\ref{table-1}. All three models achieve strong accuracy on both tasks, especially considering that the expected accuracy from random guessing is only 33\% for YCD and 20\% for YCDLW. While these results may initially appear impressive, a deeper analysis is required to understand the underlying factors contributing to the models' performance. To better understand the models' behavior, we examine the confusion matrices for each task. Figure~\ref{figure-3} presents the confusion matrices for the YCD classification, while Figure~\ref{figure-4} shows the corresponding results for the YCDLW.

\begin{table}[h]
\centering
\caption{The results of three baseline models on the YCD and YCDLW image sets.}
\label{table-1}
\vspace{2mm}
\begin{tabular}{lccc}
    \hline
             & ConvNeXt V2 & EfficientVit & MViTv2 \\
    \hline
    YCD      & 84.21\%     & 85.01\%      & 86.53\%\\
    YCDLW    & 68.20\%     & 69.85\%      & 68.02\%\\
    \hline
\end{tabular}
\end{table}

Interestingly, despite their different architectures, the models exhibit very similar confusion patterns in both tasks. In the YCD, all models struggle most with DataComp, suggesting it is the most difficult dataset to identify, while YFCC is consistently the easiest and most accurately predicted. The same pattern holds in the YCDLW: all models perform well on YFCC but show reduced accuracy on DataComp and LAION.

\begin{figure}[h]
\centering
\includegraphics[width=0.8\columnwidth]{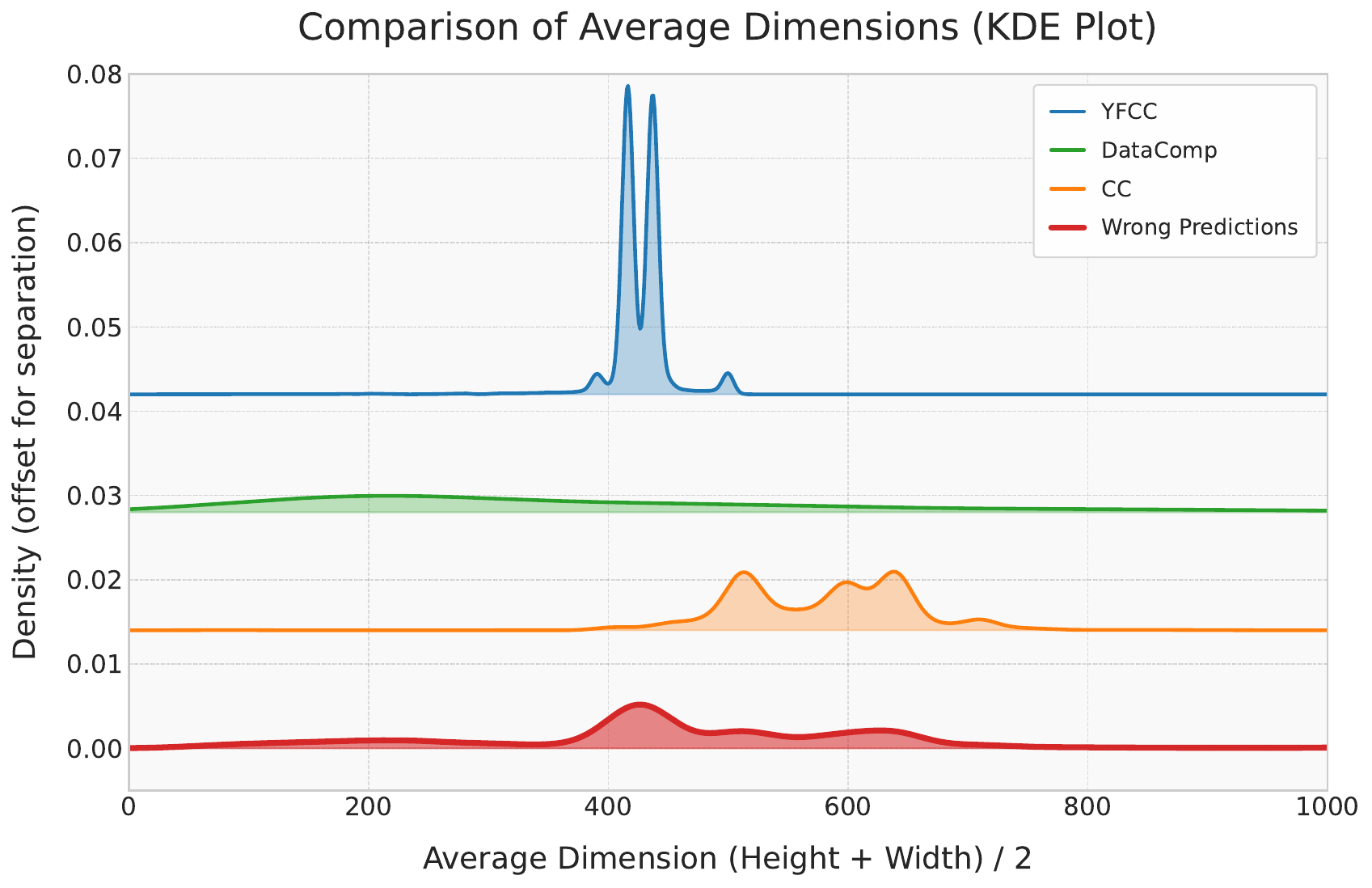} 
\caption{KDE plots of average image resolution for YFCC, CC, and DataComp datasets, and misclassified samples from three baseline models combined.}
\label{figure-5}
\end{figure}

As noted in the introduction, we observed noticeable differences in image resolution across datasets. To investigate this systematically, we analyzed the distribution of image resolutions, using the average of each image's height and width as a one-dimensional proxy. We then applied KDE to visualize these distributions. For YCD, Figure~\ref{figure-5} shows the resolution distributions for the YFCC, CC, and DataComp datasets, along with samples misclassified by all three baseline models combined. The plot reveals that most misclassified images fall within regions where resolution distributions overlap. This pattern suggests that the models may rely, at least in part, on resolution as a shortcut. When resolution cues become ambiguous due to overlap between datasets, misclassification appears more likely. While this does not confirm resolution as the sole driving factor, it highlights a key limitation in using dataset classification accuracy as evidence of meaningful semantic divergence between datasets.

\begin{figure}[h]
\centering
\includegraphics[width=0.8\columnwidth]{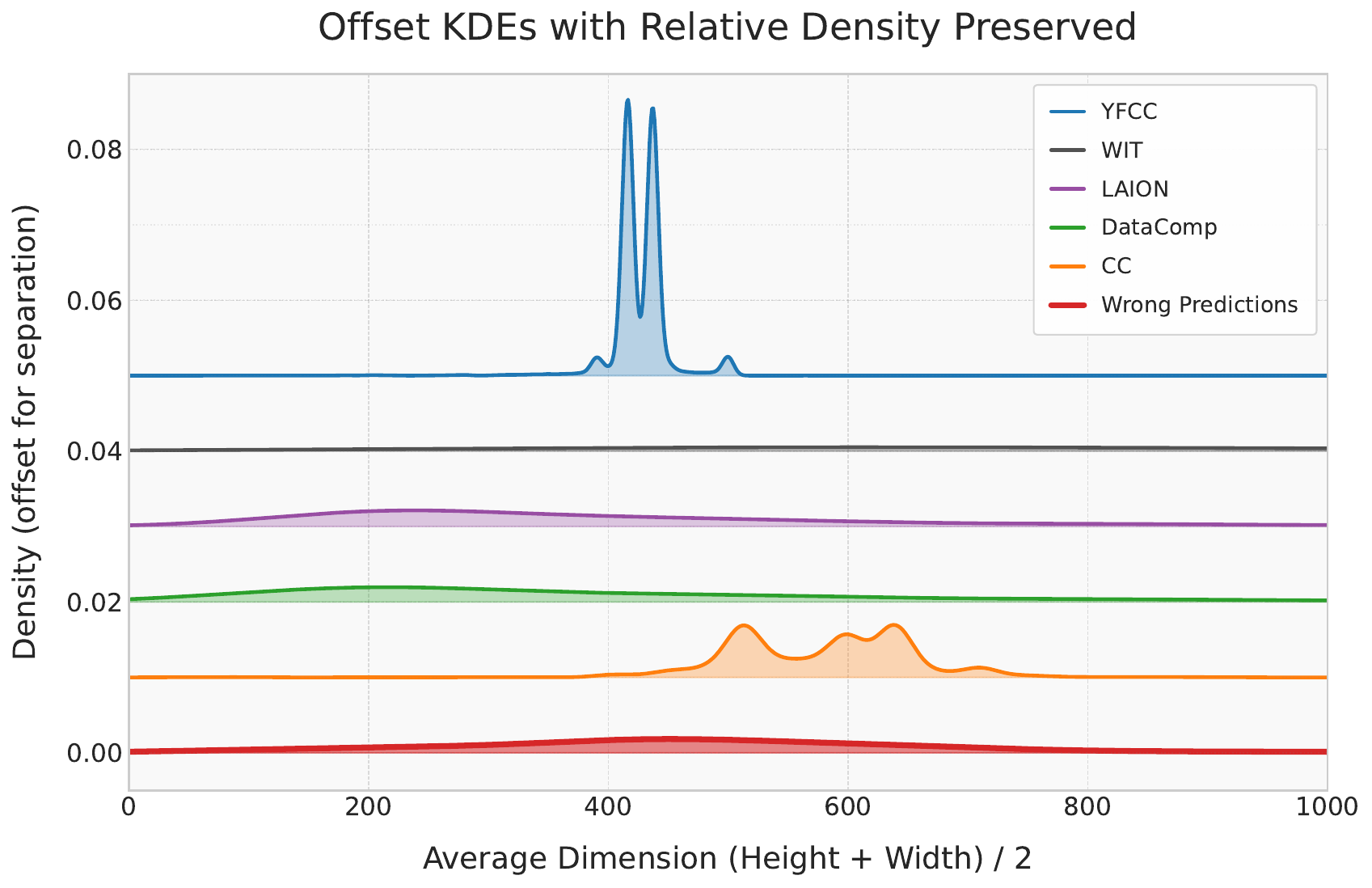} 
\caption{KDE plots of average image resolution for YFCC, CC, DataComp, WIT, and LAION datasets, along with the resolution distribution of misclassified samples aggregated across all three baseline models.}
\label{figure-6}
\end{figure}

Similarly, for YCDLW, Figure~\ref{figure-6} shows the resolution distributions of all five datasets alongside misclassified samples aggregated across the three models. Once again, misclassification patterns align closely with regions where resolution distributions overlap, reinforcing the idea that image resolution plays a dominant role in model predictions—even in more complex classification settings. Notably, DataComp and LAION share very similar resolution profiles, likely due to their common origin in Common Crawl and similar curation methods, including filtering images to match captions within the CLIP \citep{radford2021clip} embedding space. Also, WIT exhibits a long-tailed but much narrower resolution distribution, primarily concentrated above 500 pixels and extending into the thousands, which results in less overlap with other datasets and may reduce confusion.

Across all architectures we tested, consistent patterns emerge: models perform best on YFCC, which has a tightly clustered resolution distribution, and struggle most with DataComp and LAION, which have broad, overlapping resolutions. These results persist despite significant differences in model design, strongly indicating that resolution is probably the primary factor driving classification performance.

\subsection{Designed Experiments}

So far, we have demonstrated that resolution-induced artifacts likely play a significant role in dataset classification. However, to move beyond observational correlations and obtain stronger evidence, we designed targeted experiments. Our goal is to systematically validate that resolution is not merely correlated with performance but is, in fact, a primary driver. For all these experiments, we use the YCD combination as the target image set for dataset classification.

\subsubsection{Fake Images}
\begin{figure}[h]
\centering
\includegraphics[width=0.7\columnwidth]{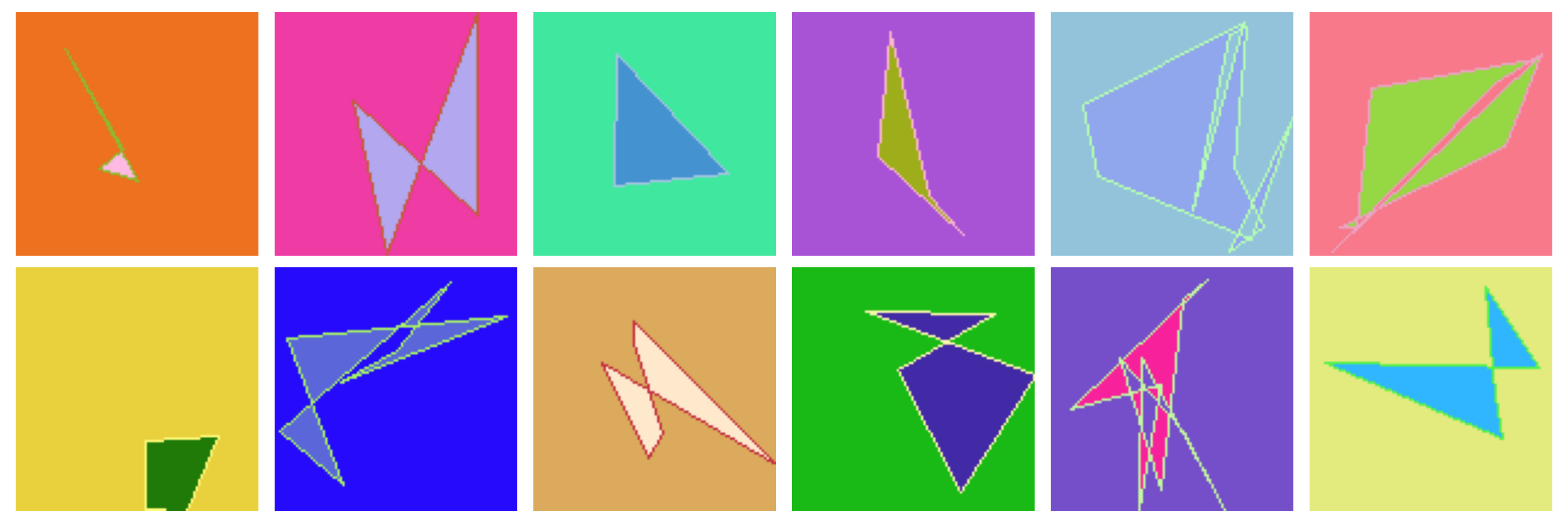} 
\caption{Several samples of generated fake images at 100×100 resolution.}
\label{figure-7}
\end{figure}

\begin{figure}[t]
\centering
\includegraphics[width=0.7\columnwidth]{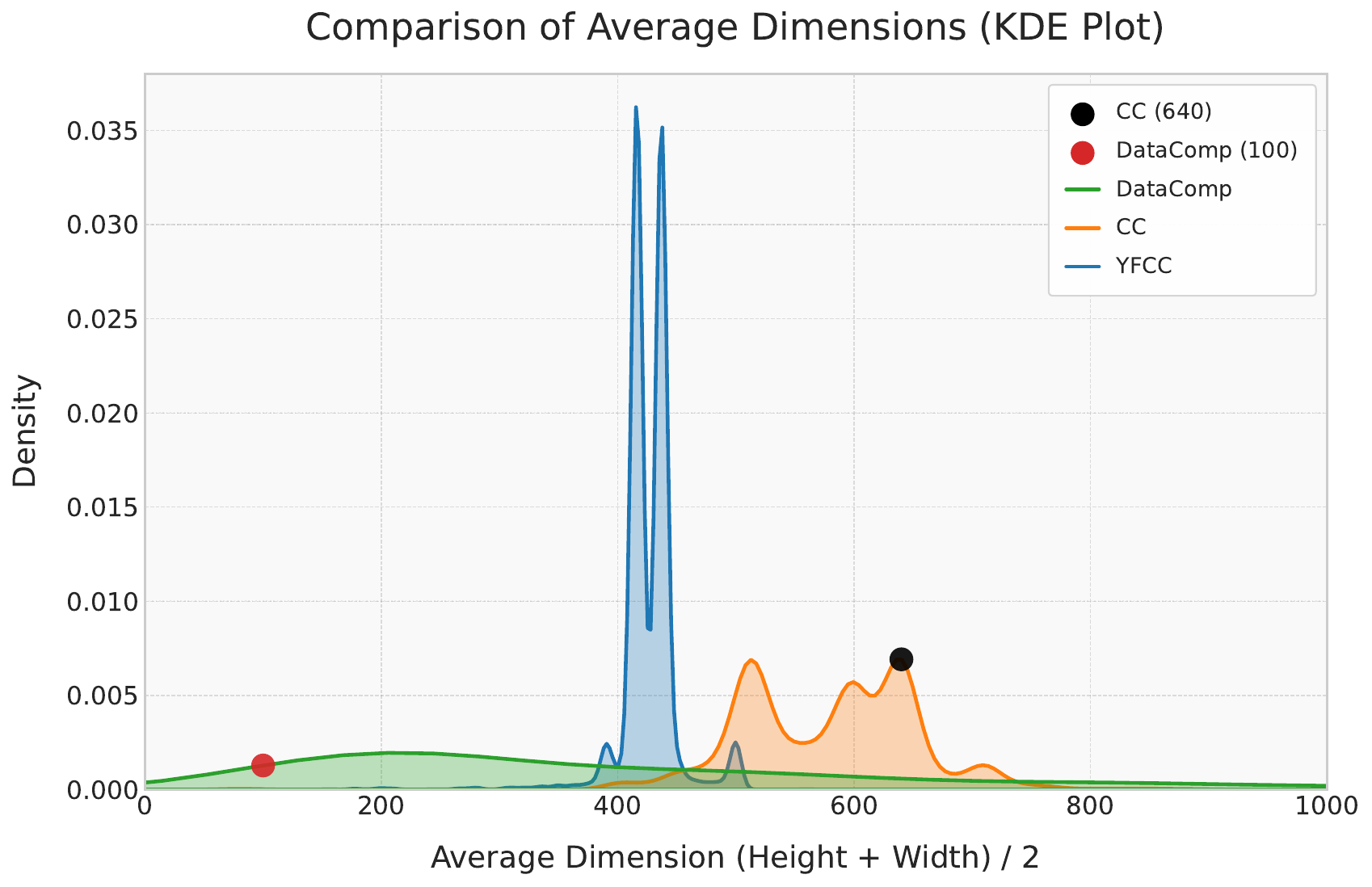} 
\caption{Distribution of average image resolutions across datasets, with DataComp dominating at resolution 100 and CC at resolution 640.}
\label{figure-8}
\end{figure}

To isolate the influence of image resolution from semantic content, we constructed a set of entirely synthetic \emph{``fake''} images comprising non-semantic, procedurally generated textures, created across a range of fixed resolutions. Examples of fake images at resolution 100×100 are shown in Figure~\ref{figure-7}. These images were generated to match the resolution distributions observed in real datasets while intentionally excluding any semantic information. Focusing on the resolution distribution of the YCD image set, we selected two representative resolution points: 100 and 640. As you can see in Figure~\ref{figure-8}, at resolution 100, DataComp dominates the distribution, whereas at 640, CC is more prevalent.

When we fed synthetic images into models trained for the YCD dataset classification task, we expected random predictions due to the absence of meaningful content. However, as shown in Table~\ref{table-2}, the results tell a different story. Among 1,000 fake images sized $100\times100$ pixels, a resolution common in DataComp, the vast majority were classified as DataComp by all three baseline models. Similarly, for synthetic images sized 640×640 pixels, which is typical for CC, the majority were classified as belonging to CC. These results demonstrate that the models rely heavily on resolution-based cues alone, even when semantic content is entirely absent. This reinforces the conclusion that dataset classification is strongly influenced by superficial, low-level artifacts rather than genuine semantic understanding.

\begin{table}[h]
\centering
\caption{Predictions for YFCC, CC, and DataComp at two resolutions (100 and 640) for each model.}
\label{table-2}
\vspace{2mm}
\setlength{\tabcolsep}{1.5mm}
\begin{tabular}{lcccc}
    \hline
    Model & Resolution & YFCC & CC & DataComp \\
    \hline
    \multirow{2}{*}{ConvNeXt V2} & 100 & 80 & 112 & 808 \\
                                 & 640 & 264 & 510 & 226 \\
    \hline
    \multirow{2}{*}{MViTv2}      & 100 & 153 & 101 & 746 \\
                                 & 640 & 247 & 493 & 260 \\
    \hline
    \multirow{2}{*}{EfficientVit}& 100 & 102 & 67  & 831 \\
                                 & 640 & 196 & 571 & 233 \\
    \hline
\end{tabular}
\end{table}

\subsubsection{Two-Step Resizing}
To rule out the resizing artifacts which are serving as classification cues, we modified the pre-processing pipeline by introducing an intermediate resizing step. Instead of directly resizing images to 224×224, we first resized them to 112×112 and then upsampled them to 224×224. This two-step resizing process aimed to reduce the impact of dataset-specific resizing artifacts.

The results, as demonstrated in Table~\ref{table-3}, revealed a noticeable drop in model performance, indicating that the models had previously relied on these artifacts as part of their decision-making process. Although this approach did not entirely eliminate the artifacts, it significantly reduced their influence. This suggests that low-level resizing effects can play a nontrivial role in dataset classification tasks.

\begin{table}[h]
\centering
\caption{Performance of three baseline models on YCD with Normal training vs Two-step resizing.}
\label{table-3}
\vspace{2mm}
\begin{tabular}{lcc}
    \hline
    Model & Normal Training & Two-Step Resizing \\
    \hline
    ConvNeXt V2 & 84.21\% & 70.91\% \\
    MViTv2      & 86.53\% & 68.36\% \\
    EfficientVit& 85.01\% & 71.64\% \\
    \hline
\end{tabular}
\end{table}

\subsubsection{Residual Images}
\begin{figure}[h]
\centering
\includegraphics[width=0.5\columnwidth]{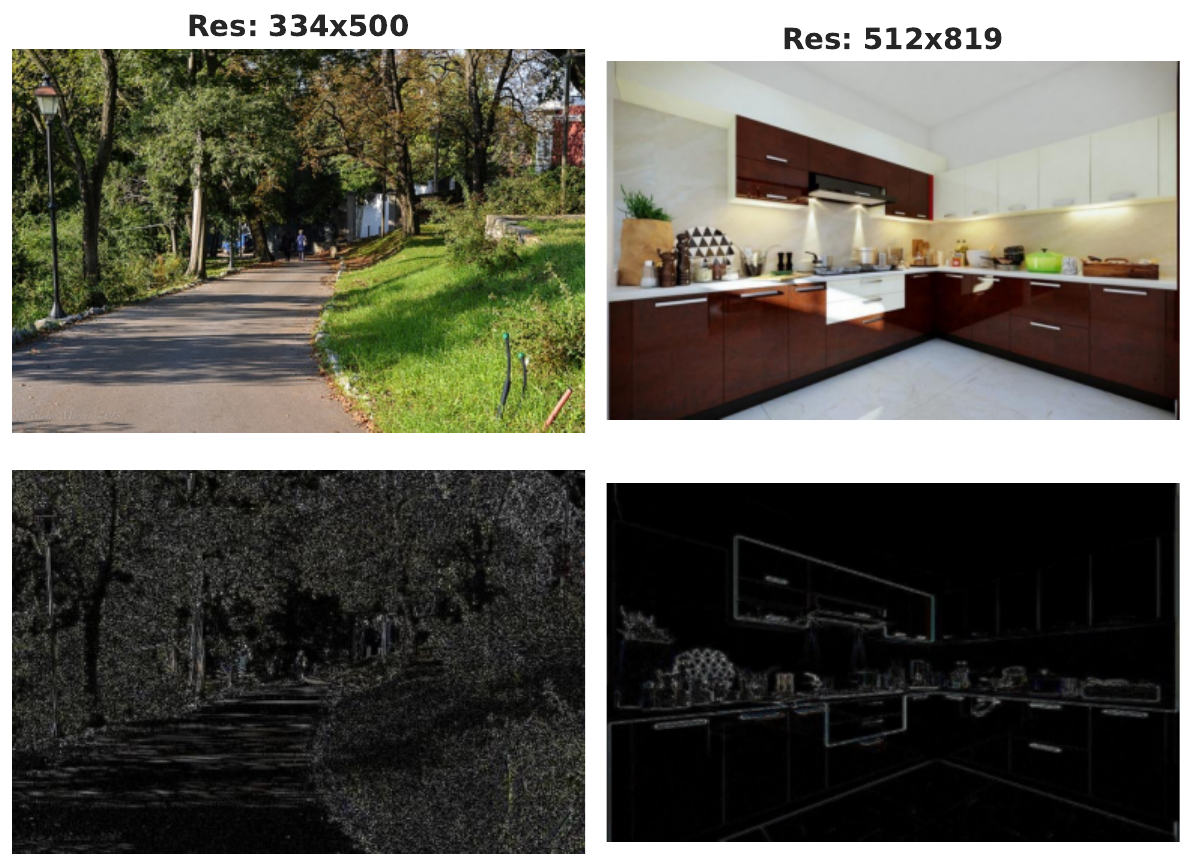} 
\caption{Residual images highlighting artifacts introduced by resizing.}
\label{figure-9}
\end{figure}

To directly isolate and visualize the artifacts introduced by standard resizing operations, we introduce the concept of residual images. For each image, we first downsample it to 224×224 and then upsample it back to its original resolution. By subtracting the original image from its resized version, we obtain a residual image that captures the distortions and cues introduced purely by resizing. Examples of these residual images are presented in Figure~\ref{figure-9}.

To assess whether these artifacts alone are sufficient for dataset classification, we trained baseline models using only the residual images derived from the YCD image set. Surprisingly, as shown in Table~\ref{table-4}, the models not only retained strong classification performance but, in all cases, outperformed those trained on the original images. This finding suggests that the resizing process introduces dataset-specific signals that models can exploit. These signals do not appear to be random noise, but rather function as distinctive dataset fingerprints.

\begin{table}[h]
\centering
\caption{Classification performance of baseline models on normal vs. residual images.}
\label{table-4}
\vspace{2mm}
\begin{tabular}{lcc}
    \hline
    Model & Normal Images & Residual Images \\
    \hline
    ConvNeXt V2 & 84.21\% & 88.46\% \\
    MViTv2      & 86.53\% & 89.37\% \\
    EfficientVit& 85.01\% & 88.87\% \\
    \hline
\end{tabular}
\end{table}

\subsubsection{Number of training samples}

\citet{liu2024decade} observed that dataset classification accuracy improves with more training samples. They interpreted this as evidence that models learn generalizable semantic patterns rather than merely memorizing examples. However, when viewed through the lens of resolution-based bias, this phenomenon admits an alternative explanation: as the number of training samples increases, the model receives a more faithful estimate of the dataset's resolution distribution, which dominates the classification signal.

\begin{figure}[h]
\centering
\includegraphics[width=0.6\columnwidth]{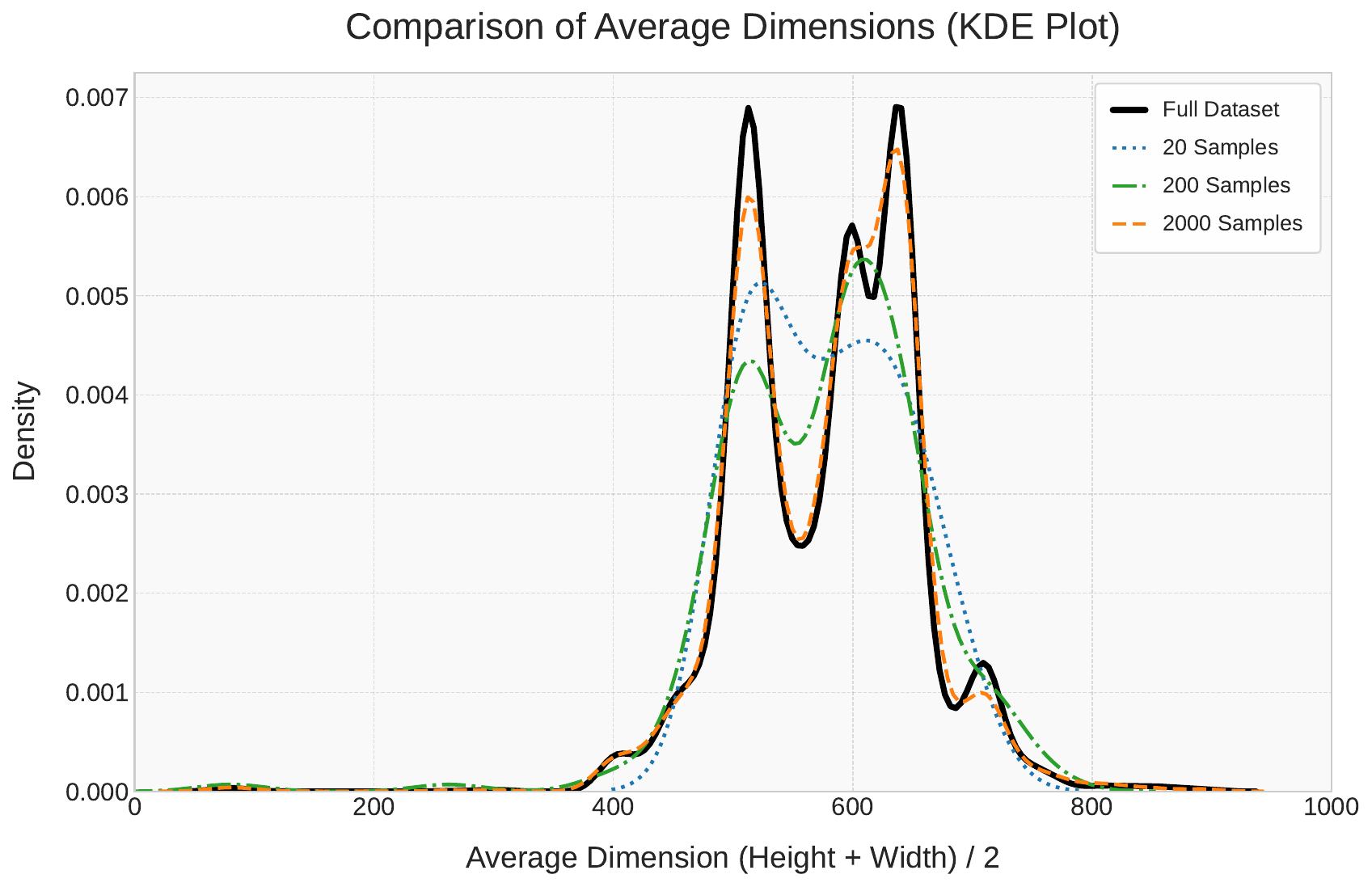} 
\caption{KDE plots of the average image resolution in the CC dataset using different sample sizes (20, 200, 2000, and full dataset).}
\label{figure-10}
\end{figure}

To support this view, Figure~\ref{figure-10} displays KDE plots of the CC dataset's resolution distribution at different sample sizes. As shown, even small samples approximate the global shape, but larger subsets (e.g., 2000 samples) yield near-perfect matches to the full distribution. This reinforces our core hypothesis: improved classification with more training data arises not from deeper semantic understanding, but from finer modeling of resolution statistics.

\section{Experiments: Unsupervised Semantic Bias Assessment}
\label{sec5}

Having shown that supervised dataset classification is fundamentally compromised by its reliance on low-level signals like resolution, we now evaluate our unsupervised framework as a principled alternative. We conduct this evaluation on the web-scale natural image datasets (YCD, YCDLW) that are the focus of this work, as they represent the primary testbed for the \emph{``Name That Dataset''} paradigm we critique. We systematically assess whether it provides a more reliable measure of semantic bias by applying our clustering-based approach to a semantically-grounded feature space. The central question is whether the high separability reported by supervised methods persists in this space or largely disappears.

\subsection{Recovering Dataset Identity with a Pretrained Model Without Supervision}

As discussed before, our proposed method assesses dataset bias by comparing clustering structure in a semantically meaningful feature space, rather than relying on supervised classification. Specifically, we extract image features using a pretrained DINOv2 model, embedding each image \(x_i\) into a high-dimensional representation \(\Phi_i = \Phi_{\mathrm{DINO}}(x_i)\). To make clustering more tractable, we reduce these features to 20 dimensions using the UMAP algorithm. This step is important for K-means clustering, which relies on feature distances that become less meaningful in high-dimensional spaces due to the curse of dimensionality. We then perform \(K\)-means clustering on the reduced features, using 100 random initializations and setting the number of clusters to match the ground-truth labels. The resulting cluster assignments are evaluated against the true labels using clustering accuracy and NMI. As demonstrated in \ref{app:robustness}, our results are robust to multiple methodological choices, including dimensionality reduction method (UMAP, PCA, or none), clustering algorithm (K-means, Agglomerative, Spectral), and the number of UMAP dimensions.

To validate our pipeline, we first apply it to established benchmarks where clear semantic categories are known to exist. As shown in Table~\ref{table-5}, the method achieves high clustering accuracy and NMI on standard tasks including scene classification (MIT-67 \citep{MIT-67}), action recognition (Stanford-40 \citep{Stanford-40}), general object classification (CIFAR-10 \citep{CIFAR}), and fine-grained classification (Oxford-IIIT Pet \citep{OxfordIIITPet}). This strong performance confirms the method's capability to recover meaningful semantic structure when it is present.

\begin{table}[h]
\centering
\caption{Clustering accuracy and NMI using DINOv2-S features followed by UMAP and k-means clustering.}
\label{table-5}
\vspace{2mm}
\begin{tabular}{lccc}
    \hline
    Dataset         & Task         & Acc(\%)       & NMI (\%) \\
    \hline  
    MIT-67          & Scene Classification         & 69.65    & 82.32 \\
    Stanford-40     & Action Recognition           & 70.89    & 75.81 \\
    CIFAR-10        & Object Classification        & 81.80    & 88.73 \\
    OxfordIIITPet   & Fine-grained Classification  & 84.08    & 91.18 \\
    \hline
\end{tabular}
\end{table}

We next apply this procedure to assess the semantic bias in the YCD and YCDLW image sets, directly comparing it against supervised \emph{``Name That Dataset''} approaches. For a fair comparison, these supervised baselines utilize the same DINOv2 backbone, implemented via linear probing on frozen features and fine-tuning the model's final stage. The results reveal a stark contrast. As shown in Table~\ref{table-6}, supervised methods achieve high accuracy, whereas our unsupervised clustering yields near-chance accuracy and very low NMI. This indicates that the images do not form intrinsically separable clusters in a semantically-grounded feature space.

\begin{table}[h]
\centering
\caption{Clustering accuracy and NMI using DINOv2-S features followed by UMAP and \(K\)-means clustering on the YCD and YCDLW image sets. NMI is only applicable to unsupervised clustering. Accuracy for supervised models reflects direct label prediction; for clustering, it is computed via optimal label matching.}
\label{table-6}
\vspace{2mm}
\setlength{\tabcolsep}{4mm}
\begin{tabular}{l c c}
    \hline
    \multicolumn{3}{c}{\textbf{YCD}} \\
    \hline
    Method                   & Acc (\%) & NMI (\%) \\
    \hline
    Random Chance            & 33.33    & N/A      \\
    DINOv2 + Fine-tuning     & 85.67    & N/A      \\
    DINOv2 + Linear Probing  & 66.91    & N/A      \\
    \hline
    \textbf{DINOv2 (Ours)}   & \textbf{46.95} & \textbf{6.60} \\
    \hline
    \hline
    \multicolumn{3}{c}{\textbf{YCDLW}} \\
    \hline
    Method                  & Acc (\%) & NMI (\%) \\
    \hline
    Random Chance            & 20.00    & N/A      \\
    DINOv2 + Fine-tuning     & 70.36    & N/A      \\
    DINOv2 + Linear Probing  & 48.89    & N/A      \\
    \hline
    \textbf{DINOv2 (Ours)}   & \textbf{31.47} & \textbf{6.32} \\
    \hline
\end{tabular}
\end{table}


These results provide a definitive explanation for the puzzle we identified in the introduction, where a minimal model drastically outperformed humans on the YCD classification task in the user study performed by \citet{liu2024decade}. Our unsupervised method achieves 46.95\% accuracy on YCD, which is close to the human baseline of 45.4\%. However, we do not attribute special significance to this numerical proximity; it likely reflects chance or different failure modes rather than shared semantic understanding. The key finding is that unlike supervised methods (84-87\% accuracy), our unsupervised approach yields near-random performance, indicating that the high separability reported by supervised methods largely vanishes when dataset labels are removed from the analysis.

One might question whether this low clustering accuracy simply reflects our choice of k=3 clusters, potentially missing finer-grained semantic structure aligned with dataset boundaries. To investigate this, we performed multi-granularity clustering on the YCD image set with $k$ ranging from 2 to 10 (see \ref{app:granularity} for full results). If datasets contained distinctive semantic subcategories, NMI would be expected to increase at some k>3 as these subcategories are discovered. Instead, as detailed in \ref{app:granularity}, NMI remains consistently low across all values of k, with no upward trend as cluster count increases. This confirms that the low semantic separability we observe is not an artifact of choosing k=3.

\subsection{Analysis Across Model Scales and Pretraining Objectives}

A potential counter-argument to our low clustering scores is that the feature extractor itself may lack the capacity to discern the subtle semantic patterns necessary to distinguish between datasets. If true, our results could be a reflection of model weakness rather than genuine semantic overlap. To definitively rule out this \emph{``capacity hypothesis,''} we evaluated our pipeline across three scales of DINOv2: ViT-S/14 (small), ViT-B/14 (base), and ViT-L/14 (large). A critical aspect of this test is that it must be two-sided: if larger models are the solution, we should see a significant improvement in clustering performance on the web-scale YCD and YCDLW sets. Simultaneously, as a necessary sanity check, we must verify that these more powerful models are indeed capable of identifying finer semantic structures, which should be reflected in improved scores on established benchmarks (e.g., CIFAR-10, Oxford-Pet).

The results, presented in Tables~\ref{table-7} and~\ref{table-8}, provide a clear and unambiguous outcome. For the YCD and YCDLW datasets, clustering accuracy and NMI remain stubbornly low across all model scales. This flat trend decisively refutes the capacity hypothesis, demonstrating that even a highly capable model like ViT-L/14 cannot find a semantic clustering that aligns with dataset labels. In stark contrast, the performance on curated benchmarks like CIFAR-10 shows a substantial and consistent increase in both metrics with larger models. This divergence confirms that our feature extractors are functioning as intended: they learn more nuanced representations when greater semantic structure exists, which validates our method.

\begin{table}[h]
\centering
\caption{Clustering accuracy (\%) across DINOv2 variants and datasets.}
\label{table-7}
\vspace{2mm}
\begin{tabular}{l c c c}
    \hline
    Dataset & DINOv2-S & DINOv2-B & DINOv2-L \\
    \hline
    YCD           & 46.95 & 46.40 & 45.99 \\
    YCDLW         & 31.47 & 31.35 & 29.87 \\
    \hline
    \hline
    MIT-67        & 69.65 & 74.50 & 81.79 \\
    Stanford-40   & 70.89 & 76.26 & 76.88 \\
    CIFAR-10      & 81.80 & 92.92 & 96.43 \\
    Oxford-Pet    & 84.08 & 85.37 & 88.67 \\
    \hline
\end{tabular}
\end{table}

\begin{table}[h]
\centering
\caption{NMI (\%) across DINOv2 variants and datasets.}
\label{table-8}
\vspace{2mm}
\begin{tabular}{l c c c}
    \hline
    Dataset & DINOv2-S & DINOv2-B & DINOv2-L \\
    \hline
    YCD           & 6.60 & 6.74 & 6.58    \\
    YCDLW         & 6.32 & 6.12 & 5.38    \\
    \hline
    \hline
    MIT-67        & 82.32 & 85.53 & 88.20 \\
    Stanford-40   & 75.81 & 80.15 & 82.85 \\
    CIFAR-10      & 88.73 & 96.43 & 98.12 \\
    Oxford-Pet    & 91.18 & 92.21 & 93.05 \\
    \hline
\end{tabular}
\end{table}

Furthermore, to assess the influence of the backbone architecture and the pretraining objective on the proposed framework, we compare two fundamentally distinct visual encoders: DINOv2-B, a self-supervised Vision Transformer model trained through a teacher-student distillation process, and ConvNeXtV2-B, a convolutional model trained with a masked reconstruction objective. The comparison results are summarized in Table~\ref{table-9}.

\begin{table}[h]
\centering
\caption{Comparison of clustering performance between DINOv2 and ConvNeXtV2 on YCD and YCDLW.}
\label{table-9}
\vspace{2mm}
\begin{tabular}{l c c c c}
    \hline
    \multirow{2}{*}{Dataset} & \multicolumn{2}{c}{DINOv2-B} & \multicolumn{2}{c}{ConvNeXtV2-B}\\
    \cline{2-5}
                  & Acc (\%)   & NMI (\%)   & Acc (\%)   & NMI (\%)   \\
    \hline
    YCD           & 46.40 & 6.74  & 47.34 & 7.61  \\
    YCDLW         & 31.35 & 6.12  & 29.68 & 6.18  \\
    \hline
\end{tabular}
\end{table}

The architectural comparison yields a critical insight: the low semantic separability of web-scale datasets persists regardless of model architecture or training pipeline. DINOv2 relies on self-supervised distillation, while ConvNeXtV2 is pretrained with masked autoencoding and then fine-tuned on ImageNet. Despite these divergent approaches, both models achieve similarly low scores on YCD and YCDLW (differences <1-2 percentage points), demonstrating that the lack of semantic separability is an inherent property of the data, not an artifact of a specific model family or learning approach. 

\subsection{Robustness to resolution}

\begin{figure}[!htbp]
\centering
\includegraphics[width=0.4\columnwidth]{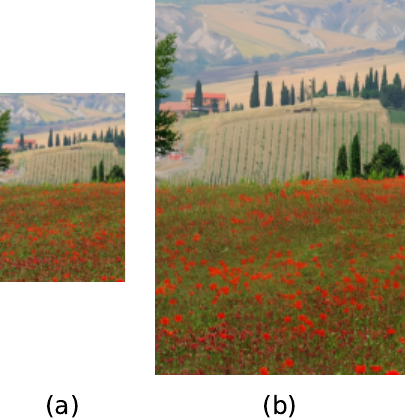} 
\caption{Example image from the DataComp dataset before and after applying super-resolution. The semantic content remains unchanged while resolution details vary.}
\label{figure-11}
\end{figure}

To assess robustness against resolution-induced cues, we upscaled the DataComp dataset images with the SwinIR \citep{SwinIR} super-resolution model. This transformation alters resolution while preserving semantic content (Figure~\ref{figure-11}). We then tasked different methods with distinguishing original from super-resolved images, despite their semantic equivalence.

The results in Table~\ref{table-10} first demonstrate that frozen DINOv2 features are not immune to low-level artifacts. If they were fully invariant to resolution cues, linear probing on these frozen features would achieve chance-level accuracy (50\%). Instead, it reaches 76.61\%, confirming that resolution artifacts persist in the feature space. Supervised methods, including fine-tuning, linear probing, and k-NN, all exploit these cues, achieving 65-95\% accuracy. This occurs even though DINOv2 was trained via self-supervised learning, a paradigm known to be less sensitive to low-level artifacts \citep{clip_traces}. The supervised objective simply forces discrimination at any cost, regardless of whether the signals are semantic or superficial. In contrast, our unsupervised clustering method has no access to dataset labels. It cannot be instructed to discriminate; it can only group images based on the natural structure of the feature space. By design, classification emerges only from what the feature space accepts as meaningful differences. Consequently, unsupervised clustering is less susceptible to non-semantic artifacts than supervised methods, achieving near-chance accuracy (50.01\%) and zero NMI on this task. This makes it a more reliable measure of true semantic bias. In other words, artifact signals present in the feature space are magnified by the supervised objective, an amplification that our unsupervised approach avoids by design.

\begin{table}[h]
\centering
\caption{
Clustering accuracy and NMI between original and super-resolved versions of the DataComp dataset.}
\label{table-10}
\vspace{2mm}
\begin{tabular}{lcc}
    \hline
    Method                  & Acc (\%)        & NMI (\%) \\
    \hline
    Random Chance           & 50.00           & N/A           \\
    DINOv2 + Fine-tuning    & 95.36           & N/A           \\
    DINOv2 + Linear Probing & 76.61           & N/A           \\
    DINOv2 + k-NN           & 65.27           & N/A           \\
    \hline
    \textbf{DINOv2 (Ours)}  & \textbf{50.01}  & \textbf{0.00} \\
    \hline
\end{tabular}
\end{table}


\section{Characterizing Semantic Bias}
\label{sec6}

\begin{figure}[t]
\centering
\includegraphics[width=0.9\columnwidth]{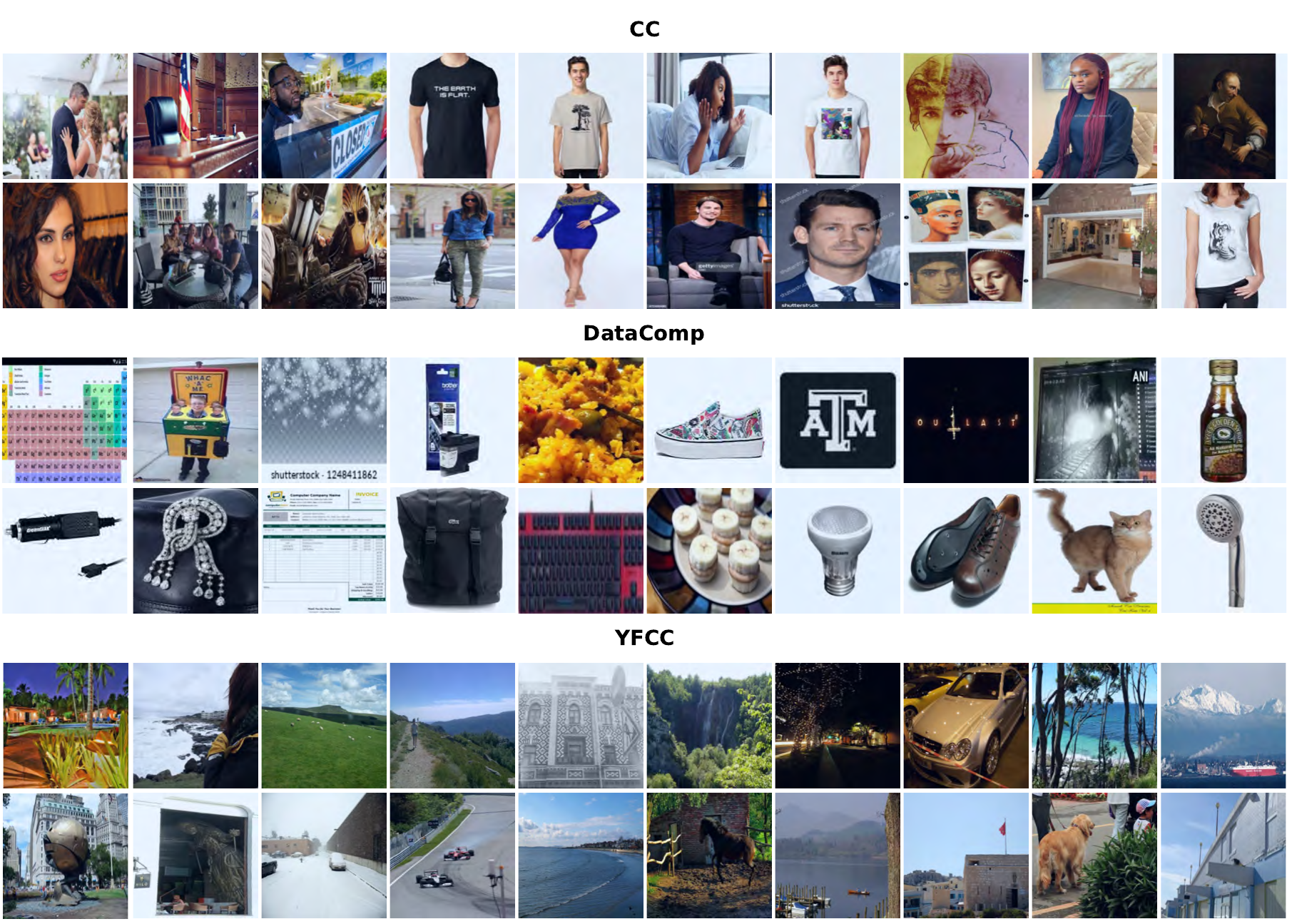} 
\caption{Visual characterization of semantic bias. Correctly clustered images reveal the dominant themes behind residual separability: CC is characterized by a \textbf{strong presence of people and indoor images}, DataComp by \textbf{commercial, object-centric} photography, and YFCC by \textbf{natural and urban views}.}
\label{figure-12}
\end{figure}

To this point, we have demonstrated that dataset bias is far less pronounced than suggested by supervised classification. While our unsupervised evaluation reveals limited separability, the residual clustering accuracy of 46.95\% on the YCD combination (using DINOv2-S) sits tellingly between the 33\% random chance and the $\sim$84\% accuracy of supervised baselines. This indicates that although some semantic bias exists, it is substantially smaller than previously claimed. This finding prompts a deeper investigation: if not artifacts, what semantic clues enable this performance? We therefore shift our focus from measuring \textit{whether} bias exists to characterizing \textit{what specific patterns} drive the observed separability. To probe the nature of these differences, we randomly selected samples from correctly clustered images of each dataset (Figure~\ref{figure-12}), which reveal the semantic themes facilitating separation.

Our qualitative examination reveals distinct semantic themes that align with each dataset's origin and collection methodology. CC images predominantly feature human subjects, indoor scenes, and clothing photography (Clothing both displayed alone and worn by models). In contrast, the images in DataComp are often composed like commercial stock photos, prioritizing clear, object-centric framing. Meanwhile, YFCC primarily features outdoor scenes like natural vistas, urban landscapes, and vehicles. These consistent patterns demonstrate that the residual separability stems not from low-level artifacts, but from high-level, content-based biases intrinsic to each data source. 

\begin{figure}[t]
\centering
\includegraphics[width=1.0\columnwidth]{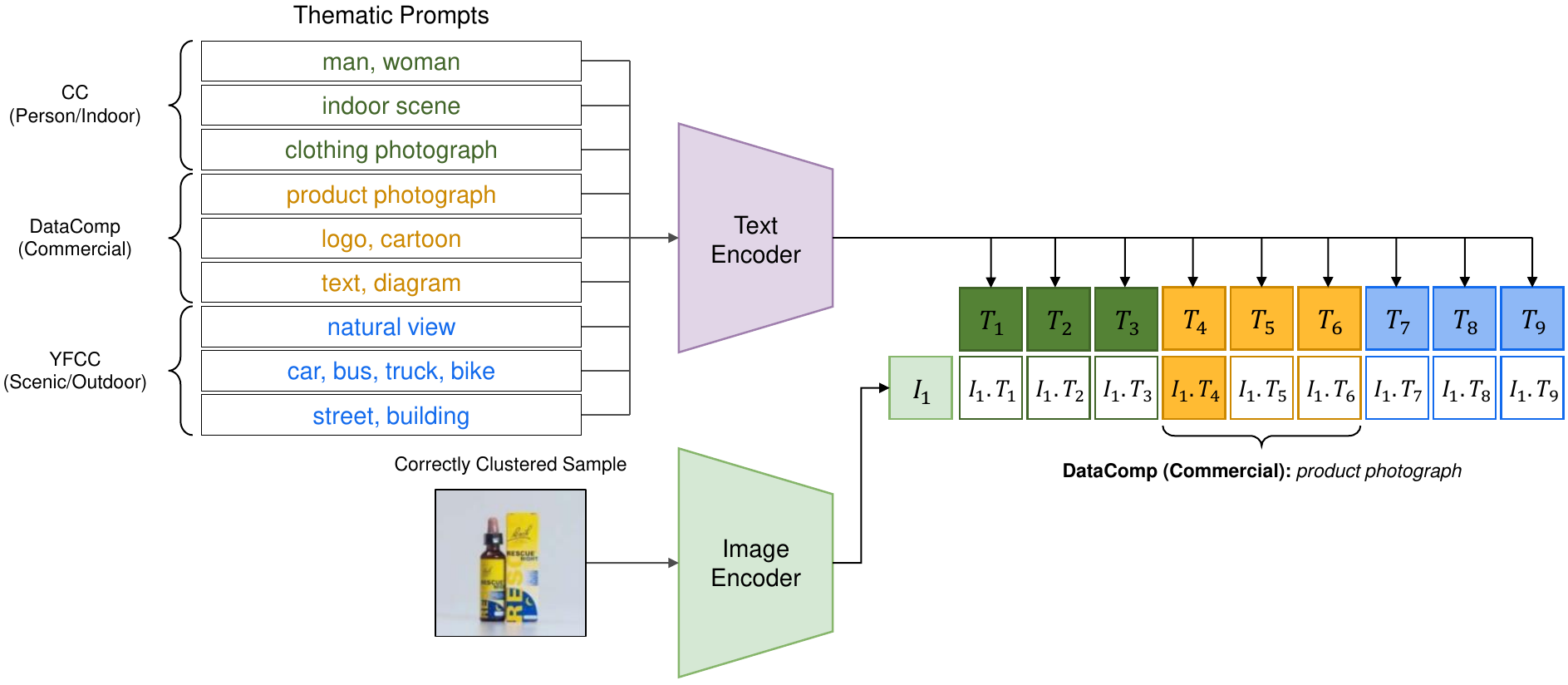} 
\caption{Pipeline for semantic characterization using CLIP. Image-prompt similarity is used to assign each correctly clustered image to a dominant semantic theme for quantitative analysis.}
\label{figure-13}
\end{figure}

To quantitatively validate these qualitative observations, we employed CLIP model with prompts directly derived from our semantic analysis. Figure~\ref{figure-13} illustrates this characterization pipeline. Based on the distinctive patterns identified through visual inspection, we designed three prompts for each dataset's characteristic domain:

\begin{itemize}
    \item \textbf{CC (Person/Indoor):} \textit{man, woman}; \textit{indoor scene}; \textit{clothing photograph}
    \item \textbf{DataComp (Commercial):} \textit{product photograph}; \textit{logo, cartoon}; \textit{text, diagram}
    \item \textbf{YFCC (Scenic/Outdoor):} \textit{natural view}; \textit{car, bus, truck, bike}; \textit{street, building}
\end{itemize}

We computed similarity scores between each correctly clustered image and all nine prompts, assigning each image to the semantic category of its single most similar prompt. As shown in Table \ref{table-11}, the resulting classification reveals a strong signal: a distinct majority of images from each dataset fall into their respective hypothesized theme, while cross-category classifications are rare. This pattern offers robust statistical evidence that the residual separability stems from genuine high-level semantic differences.

\begin{table}[h]
\centering
\caption{Quantitative semantic characterization of correctly clustered images.}
\label{table-11}
\vspace{2mm}
\begin{tabular}{l c c c c}
    \hline
    \multirow{2}{*}{Dataset} & \multicolumn{3}{c}{Semantic Category (\%)}\\
    \cline{2-4}
                 & Person/Indoor & Commercial & Scenic/Outdoor \\
    \hline
    CC           & \textbf{86.01}       & 8.50            & 5.49           \\
    DataComp     & 11.72                & \textbf{84.05}  & 4.23           \\
    YFCC         & 10.82                & 3.37            & \textbf{85.81} \\

    \hline
\end{tabular}
\end{table}

Within the framework of Section~\ref{sec3}, these semantic differences correspond to variability in the semantic component \(S\), rather than the nuisance \(N\). Although much smaller in magnitude than low-level artifacts, these shifts nonetheless reveal meaningful distributional gaps that can influence model generalization. Specifically, models pretrained on datasets that emphasize human or object-centric contexts may become biased toward those domains, potentially limiting their performance on underrepresented scene types.

Overall, our analysis demonstrates that once low-level artifacts are controlled for, residual semantic bias manifests primarily as a semantic domain bias: a reflection of the fundamental differences in visual themes, compositional styles, and subject matter intrinsic to each data source.

\section{Discussion}
\label{sec7}
\subsection{Revisiting Dataset Bias and Prior Interpretations}

The prevailing methodology for quantifying dataset bias, which trains a classifier to distinguish between datasets and interprets high accuracy as evidence of bias, rests on a fragile foundation. Our critique focuses specifically on the application of this methodology to large-scale natural image collections, which has been the primary context for the \emph{"Name That Dataset"} paradigm since its introduction. Our work systematically challenges this approach by demonstrating that what is being measured often reflects not meaningful semantic divergence, but rather superficial, low-level artifacts inherent to data collection and preprocessing pipelines. Through controlled experiments, we have established that \emph{resolution-induced artifacts} serve as a primary, robust signal for dataset classification. Models exploit these cues with high efficacy, even in the absence of semantic content (as demonstrated by our \emph{``Fake Images''} experiment) and when the signal is isolated to artifacts of the resizing process itself (as shown by our \emph{``Residual Images''} experiment).

These findings necessitate a critical re-evaluation of prior works. For instance, the recent study by \citet{liu2024decade} argued that the persistence of high classification accuracy under image corruptions (e.g., color jittering, Gaussian noise) implied models were learning features ``beyond using low-level signatures.'' We contend that this conclusion stems from an incomplete exploration of the low-level cue hypothesis. Their corruption strategies, while effective at disrupting transient artifacts like JPEG compression blocks or color statistics, are largely ineffective against the structural, resolution-based fingerprints we have identified. Consequently, the observation of resilient accuracy does not rule out low-level cues; it merely refutes a narrow subset of them.

\subsection{The ``Whac-a-Mole'' Dilemma in Supervised Diagnostics}
Our findings identify a deeper methodological issue that we term the \emph{``Whac-a-Mole Dilemma.''} In the conventional supervised setup, dataset bias is diagnosed by training a classifier to distinguish between datasets. However, this paradigm inherently incentivizes models to minimize loss by exploiting \emph{any} discriminative signal available, regardless of its semantic relevance. 

Our experiments reveal a critical pattern: when common low-level biases like JPEG compression and color quantization artifacts are neutralized through standard augmentations (color jittering, Gaussian blur), models simply pivot to alternative non-semantic cues, in our case, resolution artifacts. This pattern suggests an unwinnable game: even if resolution artifacts were successfully mitigated, models would likely discover yet another superficial signal, such as Exchangeable Image File Format (EXIF) metadata, aspect ratios, web-scraping traces, or other dataset-specific regularities. Consequently, high classification accuracy becomes a poor indicator of genuine semantic bias, reflecting only that \emph{some} statistically separable difference exists between datasets, potentially one that is entirely superficial.

Our proposed unsupervised, representation-based pipeline offers a principled alternative to this dilemma. Rather than training a model to find any difference, we probe the intrinsic structure of data within a feature space designed for semantic meaningfulness.

\subsection{Limitations and Future Directions}

While our study provides a comprehensive framework for unsupervised dataset bias assessment, it has certain limitations that point to promising future directions. Our framework's performance depends entirely on the semantic quality and invariance properties of the chosen pretrained feature extractor. This is a fundamental constraint of our approach, as the feature space determines what similarities and differences can be captured. Also, not every specialized domain has access to robust foundation models with sufficient semantic invariance, which limits the direct applicability of our approach to such domains. While this is not a fundamental barrier for natural images, given the rapid development of ever-more-powerful foundational models such as DINOv2 and DINOv3 \citep{DinoV3}, it could pose a significant challenge for highly specialized domains like remote sensing or medical imaging. In these domains, equally robust, domain-specific foundation models may not yet be available. Recent advancements in remote sensing have explored domain adaptation for change detection \citep{da2net} and unsupervised multimodal representation learning \citep{cgsl}, suggesting potential pathways for adapting our framework to such specialized domains.

This limitation is particularly consequential for medical imaging, where dataset bias poses a critical challenge to clinical deployment. Models trained on data from one hospital or scanner vendor often fail when transferred to another institution due to distribution shifts in imaging protocols, patient demographics, or acquisition equipment \citep{medical_bias}. Measuring and mitigating such biases is essential for building reliable diagnostic tools that generalize across diverse clinical settings. However, developing foundation models for medical imaging requires not only large-scale curated datasets but also domain expertise to ensure they capture clinically relevant semantics while remaining invariant to technical artifacts. Until such models become widely available, direct application of our framework to medical imaging remains an open research direction.

Additionally, while Section~\ref{sec6} offers initial insights into the semantic themes underlying residual bias, our characterization relied on manual inspection to identify high-level domains (e.g., personal/indoor for CC, commercial for DataComp, scenic/outdoor for YFCC). Although we validated these themes quantitatively using CLIP model, this methodology cannot scale to a large number of datasets and risks missing more subtle, fine-grained concepts.

This limitation suggest another research avenue. Future work could develop automated characterization techniques; for example, using advanced vision-language models like BLIP-3 \citep{BLIP-3} to identify a dataset's distinctive visual concepts and themes in a fully unsupervised manner, enabling a more granular, concept-based bias analysis.

\section{Conclusion}
\label{sec8}
This work demonstrates that supervised dataset classification, the prevailing method for measuring dataset bias in web-scale natural image datasets, should be re-evaluated, as empirical evidence shows that high accuracy in this task stems from exploitable low-level artifacts rather than meaningful semantic differences, challenging its validity as a bias benchmark.

We introduce an unsupervised alternative that clusters representations from a foundational vision model, directly assessing semantic bias without training on superficial cues. When applied to web-scale datasets, our approach reveals significantly less inherent bias within this domain than previously reported.

These findings expose the \emph{``Whac-a-Mole Dilemma''} inherent in supervised methods, where models simply find new non-semantic cues when existing ones are suppressed. We consequently shift the core question from ``can we distinguish datasets?'' to ``are they semantically separable?''

We argue for a re-evaluation of how dataset bias is assessed, suggesting that unsupervised clustering offers a more semantically-grounded alternative to supervised classification for web-scale natural image collections.

\section*{CRediT authorship contribution statement}
\textbf{Amir Hossein Saleknia}: Conceptualization, Methodology, Software, Validation, Investigation, Writing - original Draft.  
\textbf{Mohammad Sabokrou}: Conceptualization, Methodology, Writing - review \& editing, Supervision.

\section*{Declaration of Competing Interest}
The authors declare that they have no known competing financial interests or personal relationships that could have appeared to influence the work reported in this paper.
\appendix
\section{Robustness to Methodological Choices}
\label{app:robustness}
\subsection{Robustness to UMAP Dimensionality}
\label{app:umap_dim}

To demonstrate the robustness of our unsupervised clustering results to the choice of UMAP reduced dimensionality. Tables~\ref{table-a11-acc} and~\ref{table-a12-nmi} report clustering accuracy and NMI across dimensions [20, 30, 40, 50] for DINOv2-S and DINOv2-B.

\begin{table}[h]
\centering
\caption{Clustering accuracy (\%) across UMAP reduced dimensions.}
\label{table-a11-acc}
\vspace{2mm}
\begin{tabular}{l l c c c c}
\hline
\multirow{2}{*}{Dataset} & \multirow{2}{*}{Model} & \multicolumn{4}{c}{Reduced Dimension} \\
\cline{3-6}
                          & & 20 & 30 & 40 & 50 \\
\hline
\multirow{2}{*}{YCD}      & DINOv2-S & 46.95 & 46.99 & 46.92 & 46.93 \\
                          & DINOv2-B & 46.40 & 46.43 & 46.45 & 46.47 \\
\hline
\multirow{2}{*}{YCDLW}    & DINOv2-S & 31.47 & 31.65 & 31.25 & 31.10 \\
                          & DINOv2-B & 31.35 & 31.50 & 31.15 & 30.95 \\
\hline
\hline
\multirow{2}{*}{CIFAR-10} & DINOv2-S & 81.80 & 81.74 & 81.76 & 81.86 \\
                          & DINOv2-B & 92.92 & 92.77 & 92.70 & 92.83 \\
\hline
\multirow{2}{*}{MIT-67}   & DINOv2-S & 69.65 & 71.20 & 68.06 & 69.85 \\
                          & DINOv2-B & 74.50 & 76.73 & 75.31 & 76.21 \\
\hline
\end{tabular}
\end{table}

\begin{table}[h]
\centering
\caption{NMI (\%) across UMAP reduced dimensions.}
\label{table-a12-nmi}
\vspace{2mm}
\begin{tabular}{l l c c c c}
\hline
\multirow{2}{*}{Dataset} & \multirow{2}{*}{Model} & \multicolumn{4}{c}{Reduced Dimension} \\
\cline{3-6}
                          & & 20 & 30 & 40 & 50 \\
\hline
\multirow{2}{*}{YCD}      & DINOv2-S & 6.60 & 6.64 & 6.58 & 6.58 \\
                          & DINOv2-B & 6.74 & 6.72 & 6.73 & 6.76 \\
\hline
\multirow{2}{*}{YCDLW}    & DINOv2-S & 6.32 & 6.45 & 6.25 & 6.15 \\
                          & DINOv2-B & 6.12 & 6.25 & 6.05 & 5.95 \\
\hline
\hline
\multirow{2}{*}{CIFAR-10} & DINOv2-S & 88.73 & 88.83 & 88.84 & 88.62 \\
                          & DINOv2-B & 96.43 & 92.51 & 92.54 & 92.41 \\
\hline
\multirow{2}{*}{MIT-67}   & DINOv2-S & 82.32 & 82.91 & 81.83 & 82.38 \\
                          & DINOv2-B & 85.53 & 85.81 & 85.32 & 85.57 \\
\hline
\end{tabular}
\end{table}

These results reveal a consistent and telling pattern across all tested dimensionalities. The web-scale datasets, YCD and YCDLW, persistently exhibit low clustering accuracy, approximately 30-47\%, and NMI, approximately 5-7\%, which robustly confirms our central finding of their minimal intrinsic semantic separability. This stands in stark contrast to the curated benchmarks, CIFAR-10 and MIT-67, which consistently achieve high scores on both metrics, thereby validating our method's capability to correctly identify and quantify genuine semantic structure where it exists. Crucially, the performance metrics for all datasets show only minimal fluctuations, typically less than 2 percentage points, across the range of UMAP dimensions. This stability demonstrates that the profound gap in separability between web-scale and curated datasets is a fundamental property of the data itself, and that the conclusions of our study are not sensitive to this specific hyperparameter choice.

\subsection{Robustness to Dimensionality Reduction Choices}
\label{app:dim_reduction}

To verify that our findings are not artifacts of our specific dimensionality reduction method, we evaluated two alternatives to UMAP: (1) PCA as a linear reduction method, and (2) no dimensionality reduction (clustering directly on raw DINOv2-S features). Table~\ref{tab:dim_robustness} reports clustering accuracy and NMI on YCD and CIFAR-10 across these settings.

\begin{table}[h]
\centering
\caption{Clustering performance (\%) with different dimensionality reduction strategies using DINOv2-S features.}
\label{tab:dim_robustness}
\vspace{2mm}
\begin{tabular}{l c c c c}
\hline
\multirow{2}{*}{Setting} & \multicolumn{2}{c}{YCD} & \multicolumn{2}{c}{CIFAR-10} \\
\cline{2-5}
 & Acc & NMI & Acc & NMI \\
\hline
UMAP (20 dim) - our standard & 46.95 & 6.60 & 81.80 & 88.73 \\
PCA (20 dim) & 39.96 & 4.17 & 75.61 & 73.92 \\
No reduction (raw features) & 40.02 & 4.19 & 75.10 & 74.07 \\
\hline
\end{tabular}
\end{table}

The results reveal two key findings. First, on YCD, all three settings produce near-random performance (39-47\% accuracy, 4-6\% NMI). Critically, clustering on raw features achieves only 40.02\% accuracy—nearly identical to PCA and only marginally lower than UMAP. This confirms that the lack of semantic separability is intrinsic to the data itself, not an artifact of dimensionality reduction.

Second, on CIFAR-10, all methods successfully recover semantic structure (75-82\% accuracy, 73-88\% NMI), validating that our pipeline can identify genuine semantic clusters when they exist. UMAP performs best, improving accuracy from 75.10\% (raw features) to 81.80\%, suggesting it preserves semantic relationships more effectively than PCA. On YCD, UMAP similarly improves accuracy from 40.02\% to 46.95\%. The fact that UMAP improves performance on both datasets, one with strong semantic structure (CIFAR-10) and one with minimal semantic differences across datasets (YCD), demonstrates that UMAP faithfully preserves existing semantic relationships without inventing false structure.

\subsection{Robustness to Clustering Algorithm Choices}
\label{app:clustering}

To assess sensitivity to the choice of clustering algorithm, we replaced K-means with two alternative approaches on UMAP-reduced features (dim=20): Agglomerative Clustering (Ward linkage) and Spectral Clustering. Table~\ref{tab:clustering_robustness} reports the results on YCD and CIFAR-10.

\begin{table}[h]
\centering
\caption{Clustering performance (\%) with different clustering algorithms using UMAP-reduced DINOv2-S features.}
\label{tab:clustering_robustness}
\vspace{2mm}
\begin{tabular}{l c c c c}
\hline
\multirow{2}{*}{Algorithm} & \multicolumn{2}{c}{YCD} & \multicolumn{2}{c}{CIFAR-10} \\
\cline{2-5}
 & Acc & NMI & Acc & NMI \\
\hline
K-means (our standard) & 46.95 & 6.60 & 81.80 & 88.73 \\
Agglomerative Clustering & 44.49 & 4.91 & 84.48 & 88.22 \\
Spectral Clustering & 47.06 & 7.81 & 71.18 & 77.80 \\
\hline
\end{tabular}
\end{table}

All three algorithms produce near-random performance on YCD (44-47\% accuracy, 4-8\% NMI), with no method achieving meaningful separation. In contrast, all algorithms successfully recover semantic structure on CIFAR-10 (71-84\% accuracy, 77-88\% NMI), with K-means providing the best balance of accuracy and NMI. This demonstrates that the absence of clear dataset-specific clusters in web-scale datasets is robust to the choice of clustering algorithm.

\section{Multi-Granularity Clustering Analysis}
\label{app:granularity}
To investigate whether semantic differences between datasets might emerge at finer or coarser granularities than k=3, we performed k-means clustering on the YCD dataset with k ranging from 2 to 10 and evaluated alignment with dataset labels using NMI.

\begin{table}[h]
\centering
\caption{NMI (\%) for YCD image sets with varying number of clusters using DINOv2-S features.}
\label{table-b1-nmi}
\vspace{2mm}
\begin{tabular}{c*{9}{c}}
\hline
k & 2 & 3 & 4 & 5 & 6 & 7 & 8 & 9 & 10 \\
\hline
NMI (\%) & 9.02 & 6.60 & 6.29 & 6.50 & 6.34 & 6.37 & 6.54 & 6.45 & 7.43 \\
\hline
\end{tabular}
\end{table}

The results reveal two key findings. First, the highest NMI (9.02\%) occurs at k=2, indicating that even with only two clusters, alignment with three dataset labels is weak. Second, and more importantly, from k=3 to k=10, NMI remains remarkably stable, showing no upward trend as cluster count increases.

If semantic differences existed but required more clusters (e.g., if each dataset contained multiple distinct subcategories), we would expect NMI to increase substantially at some k > 3. Instead, the flat trajectory across all k values demonstrates that the low separability observed in Section 5 is not an artifact of choosing k=3, but rather reflects genuine semantic overlap between datasets.

\section*{Data Availability}
Data and code will be made available upon reasonable request.

\section*{Declaration of generative AI and AI-assisted technologies in the writing process}
During the preparation of this work, the authors used LLMs such as ChatGPT (OpenAI) to improve language readability and for editorial assistance with manuscript structure. After using this tool, the authors reviewed and edited the content carefully and take full responsibility for the content of the published article.


\bibliographystyle{elsarticle-num-names} 
\bibliography{cas-refs}






\end{document}